\documentclass[preprint,12pt,authoryear]{elsarticle}

\usepackage{setspace}
\usepackage{amssymb}

\usepackage{tikz}
\usepackage{subcaption}
\usepackage{hyperref}
\usepackage[capitalize]{cleveref}
\usepackage{pgffor}
\usepackage{booktabs}
\usepackage{multicol}
\usepackage{multirow}
\usepackage{natbib}
\usepackage[tight-spacing=true]{siunitx}
\DeclareSIUnit\pixel{px}
\usepackage{csquotes}
\usepackage[super]{nth}
\usepackage{pgfplots}
\usepackage{epflcolors}
\pgfplotsset{compat=1.16}

\usepackage{helvet}

\newcommand{\classname}[1]{{\textsl{#1}}}
\newcommand{\metric}[1]{{\textsc{#1}}}
\newcommand{\regionname}[1]{{\textsl{#1}}}
\newcommand{\modelname}[1]{{\textsc{#1}}}
\newcommand{\new}[1]{{#1}}

\usepackage{lineno}

\journal{ArXiv}

\begin{document}

\begin{frontmatter}



\title{Large-scale Detection of Marine Debris in Coastal Areas with Sentinel-2}


\author{Marc Ru\ss{}wurm, Sushen Jilla Venkatesa, Devis Tuia}

\affiliation{organization={EPFL ECEO Laboratory},
            addressline={Rue de l'Industrie 17}, 
            city={Sion},
            postcode={1950}, 
            state={Valais},
            country={Switzerland}}

\begin{abstract}
Detecting and quantifying marine pollution and macro-plastics is an increasingly pressing ecological issue that directly impacts ecology and human health. Efforts to quantify marine pollution are often conducted with sparse and expensive beach surveys, which are difficult to conduct on a large scale. Here, remote sensing can provide reliable estimates of plastic pollution by regularly monitoring and detecting marine debris in coastal areas.
Medium-resolution satellite data of coastal areas is readily available and can be leveraged to detect aggregations of marine debris containing plastic litter. 
In this work, we present a detector for marine debris built on a deep segmentation model that outputs a probability for marine debris at the pixel level. 
We train this detector with a combination of annotated datasets of marine debris and evaluate it on specifically selected test sites where \new{it is highly probable} that plastic pollution is present in the detected marine debris. 
We demonstrate quantitatively and qualitatively that a deep learning model trained on this dataset issued from multiple sources outperforms existing detection models \new{trained on previous datasets} by a large margin. \new{Our experiments show, consistent with the principles of data-centric AI, that this performance is due to our particular dataset design with extensive sampling of negative examples and label refinements rather than depending on the particular deep learning model.}
We hope to accelerate advances in the large-scale automated detection of marine debris, which is a step towards quantifying and monitoring marine litter with remote sensing at global scales, and release the model weights and training source code\footnote{\url{https://github.com/marccoru/marinedebrisdetector}}. 
\end{abstract}


\end{frontmatter}


\section{Introduction}
\label{}

Marine litter is accumulating at alarming rates, with 19 to 23 million metric tonnes dispersed in 2016 alone \citep{borrelle2020predicted}. Plastic artifacts constitute 75\% of marine litter, exceeding 5 trillion objects in numbers \citep{eriksen2014plastic}, and are causing a serious threat to marine ecosystems and human health. 
Approximately 80\% of marine litter originates from terrestrial sources \citep{andrady2011microplastics}. It accumulates in rivers \citep{van2019seine,van2020plastic} and lakes \citep{alencastro2012pollution} and eventually enters open oceans.
\new{Primary micro-plastics are purposefully manufactured to carry out a specific function, like abrasive particles or powders for injection molding. Secondary micro-plastics result from fragmentation of larger objects \citep{GESAMP}.
In particular transport in rivers} causes macro-plastics ($>$ \SI{2.5}{\centi\meter} diameter) to decompose into \new{meso- (\SI{5}{\milli\meter} $-$ \SI{2.5}{\centi\meter}) and} micro-plastics ($<$ \SI{5}{\milli\meter} diameter) \citep{GESAMP,hanke2013guidance}, which then enter the food chain.
Micro-plastics have been found across the entire planet and have been detected in antarctic penguins \citep{bessa2019microplastics}, deep-sea sediments \citep{van2013microplastic}, and human stool \citep{schwabl2019detection} and have been shown to affect the growth of corals \citep{chapron2018macro}.
A range of economic costs can also be associated with marine pollution, from clean-up expenses to loss of tourism revenue \citep{beaumont2019global}.
It is clear that monitoring and mitigating water pollution is a major environmental, social, and economic challenge, and systematic mapping is needed to both identify pollutants and measure the success of awareness and clean-up programs.
Continuous monitoring and litter quantification are often limited to individual surveys that are labor-intensive and expensive to conduct regularly \citep{van2016empirical}. These approaches can only cover a comparatively small area, even when surveyors are supported by aerial UAV imagery, as explored by \citet{wolf2020machine,goddijn2022using,escobar2022aerial,topouzelis2019detection}.
Effectively, only a few developed countries, such as the United Kingdom, can afford a systematic monitoring program \citep{rees1995marine}. These programs still require support from the local population in citizen science projects to collect ground data \citep{hidalgo2015contribution}. This level of engagement requires a public sensitivity to the problem, awareness, and, eventually, the technological means to report pollutants.

Satellite imagery that provides data at reasonable spatial and high temporal resolution can support this monitoring \new{in large marine areas \citep{hanke2013guidance}}. 
Even though it is a pressing issue, remote sensing-enabled monitoring of marine debris has only relatively recently emerged as a major research topic, as summarized by the broad reviews of \citet{salgado2021assessment}\new{ and \citet{TOPOUZELIS2021112675}. Both reviews} compared drone, aircraft, and optical and radar satellite-based acquisition methods.
\new{In particular, machine learning models have been increasingly used for this problem, as summarized by \citet{POLITIKOS2023106466}, who aggregated a comprehensive list of approaches and locations where machine learning algorithms have been deployed in the last years across the globe.}
For optical sensors, high spatial ($<$ \SI{3}{\meter}) and spectral resolutions beyond RGB (\SI{400}{\nano\meter} to \SI{2500}{\nano\meter}) were found optimal for the detection of aggregations of marine debris. Synthetic Aperture Radar (SAR) \new{can be potentially} suitable for detecting sea-slicks \citep{davaasuren2018detecting} that are associated with surfactants and change the surface tension of the water, which in turn reduces the radar back-scatter. These slicks consist of microbial bio-films that can be connected with micro-plastics suspended in the sea-surface microlayer \citep{salgado2021assessment}. \new{However, a recent study \citep{sun2023effects} demonstrated that only very high concentrations of microplastics lead to a sufficiently strong dampening of waves to be detectable with radar satellites.}
Similarly to sea slicks, macro-plastics can aggregate in lines driven by environmental forces, such as wind speed, waves, or coastal fronts. For instance, windrows are accumulations of surface debris. Their geometry allows for efficient ship-based collection efforts, which can be highly effective, as demonstrated by \citet{ruiz2020litter}. Their collection campaign lasted 68 working days during the spring and summer of 2018 and gathered 16.2 tons of floating marine litter in the Bay of Biscay. This work demonstrated that detecting and collecting aggregated debris on the sea surface in geographic areas with a high pollution level can be directly attributed to macro plastic litter. 
Marine debris aggregations in windrows are sufficiently large to be detectable at medium resolutions of \SI{10}{\meter} by \SI{10}{\meter} achievable by {Sentinel-2} and can effectively serve as a proxy for macro plastic litter in the oceans \citep{cozar2021,arias2021advances}.
\new{However, further distinguishing floating objects of natural origins, such as driftwood, or patches of algae and sargassum, from objects of human origins in large-scale medium-resolution imagery remains challenging and is an ongoing topic of current research \citep{HU2021112414,HU2022114082,CIAPPA2021112457,rs14102409}. This further fine-grained distinction may require currently unavailable sensor technology \citep{salgado2021assessment} and is beyond the scope of this work.}
\new{Instead, we study the effectiveness of detecting heterogeneous marine objects of both natural or anthropogenic origins at a large scale with globally available Sentinel-2 imagery.}
In this work, we aim to monitor floating marine litter by detecting marine debris as a proxy at a large scale. To do so, we evaluate our detector in selected areas where it is likely that marine litter is present in marine debris due to local studies and reports in the news and social media. This evaluation strategy ensures that our detector is sensitive to plastic pollution if marine debris is detected.
This work follows the principles of data-centric AI \citep{whang2023data}, where the methodological innovation is concentrated on carefully designing of the dataset rather than the specificities of the particular deep learning model.

\new{Throughout this work, we will use the term \emph{marine litter} according to the \citet{unep2009} definition as \emph{any persistent, manufactured, or processed solid material discarded, disposed of, or abandoned in the marine and coastal environment}. We use \emph{marine debris} more broadly as \emph{any aggregation of floating materials on the sea surface} that may or may not contain \emph{marine litter} of anthropogenic origins.
The terms \enquote{litter}, \enquote{debris} and \enquote{plastic} have particular meanings to different groups of people depending on the scientific or technical context or cultural preference \citep{GESAMP} and \enquote{marine debris} is often, especially in US-English, used synonymously with \enquote{marine litter}.
However, we believe a distinction is necessary for technical reasons in this application: visual inspection of the current satellite imagery (without on-site knowledge) can not reliably distinguish marine litter of human origins from marine debris that may also be of natural origins.
Hence, any work relying on hand annotations of satellite images can not resolve this conflict objectively, as on-site knowledge of the composition of the visible marine debris is only available from dedicated campaigns \citep{topouzelis2019detection,topouzelis2020remote} that yield few thoroughly analyzed pixels.
In prior work \citep{mifdaletal2020towards}, we used the generic term \enquote{floating object}, while others like \citet{booth2022high} chose the term \enquote{suspected plastics}.
Both terms entail their limitations by being either too broad, as \enquote{floating objects} may include ships, or are too focused on plastics over other forms of litter.
Our definitions of anthropogenic \emph{marine litter} and generic \emph{marine debris} follow the practices of \citet{kikaki2022marida} who annotated similar objects termed marine debris in the {Marine Debris Archive} (MARIDA) and are used consistently throughout this work. 
}


The rest of the paper is organized as follows:
The next section summarizes related work on detecting marine pollution with remote sensing technology. 
\Cref{sec:materials_and_methods} describes training, validation, and evaluation data used in this study and details the implementation of the segmentation models in the Marine Debris Detector.
\Cref{sec:results} presents results compared to related work and methodologies qualitatively and quantitatively. Further experiments test the robustness of the Marine Debris Detector concerning atmospheric correction and test the transferability to higher-resolution PlanetScope imagery that can supplement the {Sentinel-2} imagery used primarily in this work. The final \cref{sec:discussion_and_conclusion} discusses the results and provides conclusions for future work.


\section{Related Work}
\label{sec:related_work}

Detecting marine debris with satellite imagery at high (typically \SI{3}{\meter} to \SI{7}{\meter} with PlanetScope imagery) and medium resolution (mainly at \SI{10}{\meter} with Sentinel-2)  is a rising scientific question in remote sensing research. Initial advances were made by pixel-wise classifiers using multi-spectral spectral reflectance in combination with dedicated spectral indices, such as the Normalized Difference Vegetation Index (NDVI). \Citet{themistocleous2020investigating} investigated the detection of floating plastic litter from space using {Sentinel-2} imagery in Cypris and proposed plastic index as the ratio of near-infrared reflectance to the sum of red and near-infred similar to NDVI.
Similarly, \citet{biermann2020finding} proposed a Floating Debris Index (FDI), which is a modification of the Floating Algae Index (FAI) \citep{hu2009novel}. They demonstrated the effectiveness of FDI with a na\"ive Bayes classifier in two-dimensional NDVI-FDI feature space. However, this classifier, originally fitted on hand-selected training and evaluation data under optimal conditions, was not accurate enough on unfiltered satellite imagery in practice, as demonstrated by \citet{mifdaletal2020towards}.
\Citet{kikaki2022marida} achieved the best accuracies with a pixel-wise random forest classifier that utilized the {Sentinel-2} reflectance bands, a range of spectral indices, and textural features. In \citet{mifdaletal2020towards}\new{, we} investigated the suitability of learned spatial features with a convolutional neural network for binary marine debris detection. While their results showed general applicability towards detecting marine debris with deep segmentation models, they identified several limitations and the sensitivity to a range of false-positive detections that made their model not employable in an automated way.
Simultaneously, \citet{nasamarinedebris} annotated RGB PlanetScope imagery with bounding boxes and trained a deep object detector on the localization of marine debris. 
Most recently, \citet{gomez2022learning} focused on detecting debris in rivers with {Sentinel-2} and tested several deep segmentation models to understand and predict floating debris accumulations.
\new{Similar to this work, \citet{booth2022high} presents a supervised U-Net classifier named MAP-Mapper which is learned on the MARIDA dataset aimed to predict the density of marine debris}.

Several public datasets were made available alongside the respective publications. Both the FloatingObjects dataset \citep{mifdaletal2020towards} and the Marine Debris Archive (MARIDA) \citep{kikaki2022marida} contain {Sentinel-2} imagery with a substantial number of hand-annotations of visually detected marine debris hand-annotated. They differ mostly in the binary (debris vs other, i.e., non-debris) and multiclass (types of debris) nature of the annotations. The NASA Marine Debris dataset \citep{nasamarinedebris} focused on 3-channel RGB PlanetScope imagery with coarse bounding box annotations.

In this paper, we extend initial work of \citet{mifdaletal2020towards}
and train a deep segmentation model on the combined datasets of FloatingObjects \citep{mifdaletal2020towards} and MARIDA \citep{kikaki2022marida}. We further use additional datasets to train our detector, which we detail in the next section.

\section{Materials and Methods}
\label{sec:materials_and_methods}

\begin{figure}
    \centering
    \includegraphics{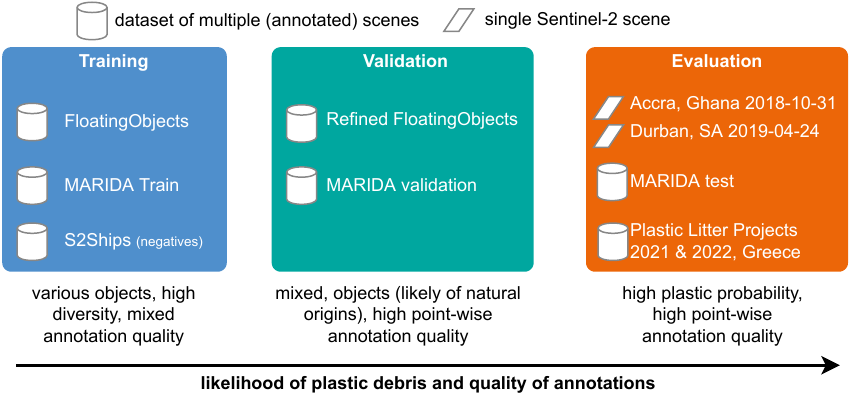}
    \caption{Overview of the datasets used for training, validation, and evaluation in this work. We focus on quantity and diversity in the training datasets while prioritizing accurate annotations in validation and evaluation data. The scenes in Accra and Durban likely contain plastic litter in the visible marine debris and are explicitly used for evaluation.}
    \label{fig:datasetfiture}
\end{figure}


Defining and aggregating training data for marine debris detection is challenging due to the heterogeneous nature of objects, the novelty of the discipline, and the scarcity of available datasets.
This section first outlines the sources, aggregation choices, and design decisions to generate the 
training, validation, and evaluation datasets used in this work. Specifically, \cref{sec:trainingdata} focuses on the datasets used for training, while \cref{sec:evaluation} outlines the validation and evaluation sets. An overview of the datasets is provided in \cref{fig:datasetfiture}.
For training datasets, we focused on quantity and aggregated a large dataset of heterogeneous marine debris and other floating materials alongside negative examples focused on ships (S2Ships). The quality of this large training pool is variable, but this also reflects the inherent difficulty of the task. In the validation and evaluation data, we focus more on the quality and accuracy of annotations of marine debris. The evaluation scenes were chosen explicitly in areas where we were certain, due to manual verification, that plastic pollution is present among marine debris.
After describing the dataset, the models used are detailed in  \cref{sec:model,sec:comparisonmodels}, which describe our detector and the comparison methods, respectively. Accuracy metrics are described in \cref{sec:metrics}.

\subsection{Training Data}
\label{sec:trainingdata}

The available annotated data on the detection of marine debris is scarce. To our knowledge, only two publicly available datasets focusing on {Sentinel-2} imagery are available today. The Marine Debris Archive (MARIDA) \citep{kikaki2022marida} provides multiple labels on polygon-wise hand-annotated {Sentinel-2} images, and the FloatingObjects \citep{mifdaletal2020towards} provides binary labels (floating objects versus water annotations) in coarse hand-drawn lines on {Sentinel-2} scenes. We further improve the quality of these annotations by an automated label refinement heuristic defined for this problem.
\new{Our goal is to train a model that can predict marine debris from openly accessible satellite imagery in different conditions and therefore making it possible process both top-of-atmosphere and atmospherically corrected bottom-of-atmosphere data. For atmospheric correction, we further chose to use products corrected with Sen2COR \citep{main2017sen2cor} that are readily available to download in Google Earth Engine rather than products corrected with ACOLITE \citep{vanhellemont2016acolite}, where the atmospheric correction would have to be done individually at each raw image scene. 
To study the effect of atmospheric correction, we test our models on imagery at different atmospheric processing levels (see \cref{sec:confusions}).}
To avoid confusion of marine debris with ships, one of the major problems highlighted in \citet{mifdaletal2020towards}, we also include the S2Ships dataset \citep{ciocarlan2021ship} that provides negative non-debris examples of class \classname{other}. All three datasets are detailed in the next subsections.

\subsubsection{FloatingObjects}
\label{sec:flobs}

The FloatingObjects dataset originates from \new{our prior work in \citet{mifdaletal2020towards}} and contains \num{26} different globally distributed {Sentinel-2} scenes. Overall, \num{3297} floating objects were annotated by lines when visually identified as \classname{marine debris}.
In this work, we use this dataset exclusively for training, as a certain level of label noise is present in the annotations. We decided to exclude four regions \regionname{accra 20181031},
\regionname{lagos 20190101},
\regionname{neworleans 20200202},
\regionname{venice 20180630}
to be re-annotated in the RefinedFloatingObjects validation dataset described later in \cref{sec:refinedFlobs}.
The remaining 22 regions
were used for training. 

We follow the data sampling strategy of \citet{mifdaletal2020towards} and crop a small image patch of \SI{128}{\pixel} by \SI{128}{\pixel} centered on each line segment of the available marine debris annotations. To obtain negative examples without any marine debris, we select random points within the {Sentinel-2} scenes and extract equally sized image patches. 
We also use both processing levels L1C (top-of-atmosphere) and L2A (bottom-of-atmosphere), where we always select the L2A image available in the Google Earth Engine Archive \citep{gorelick2017google} and resort to L1C if no atmospherically corrected image is available. The effect of atmospheric correction on the performance of the detector is evaluated later in \cref{sec:confusions}. In all cases, 12 {Sentinel-2} bands are used. These are all the available bands, excluding the haze-band B10, which the Sen2COR atmospheric correction \citep{main2017sen2cor} algorithm removes automatically. 

\begin{figure}
	\centering
	\includegraphics[width=\textwidth]{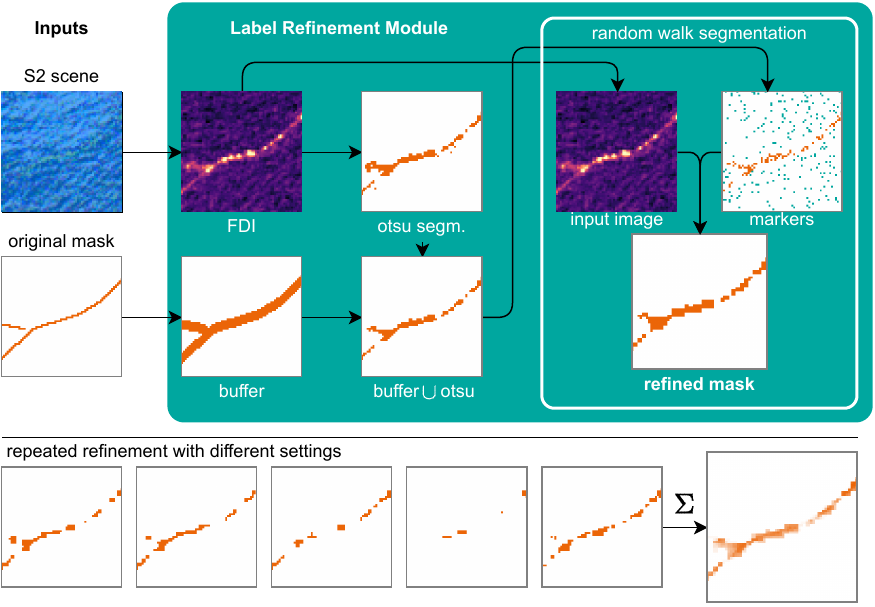}
	\caption{Label Refinement Module for the FloatingObjects dataset. It inputs a {Sentinel-2} image and the original hand annotation of the FloatingObjects dataset (left). An Otsu-threshold segmentation buffered around the hand labels \citep{otsu1979threshold} (center) is used to sample marker points (shown on the right) for a random walk segmentation algorithm \citep{grady2006random} that results in a refined annotated mask (right). By varying parameters, we generate different variants of the mask, whose average expresses the uncertainty and fuzziness on the borders of the debris (second row).}
	\label{fig:label_refinement_module}
\end{figure}

\textbf{Label Refinement Module}. While the FloatingObjects dataset provides a large number of labels, the annotated lines do not always accurately capture the width and geometry of the underlying marine debris.
We improve the hand annotations by an automated label refinement module that generates a mask that reflects more closely the geometry of the debris in the proximity of the line annotations (\cref{fig:label_refinement_module}).
The module inputs a {Sentinel-2} scene and the original line annotations mask. 
In the first stage (left side of \cref{fig:label_refinement_module}), we buffer the hand-annotated line to obtain a region of potential marine debris. Then, we calculate the Floating Debris Index (FDI) using the {Sentinel-2} scene and perform a segmentation of the FDI image with an Otsu threshold \citep{otsu1979threshold}.
The buffer and segmentation are then combined to obtain a preliminary area of marine debris in the vicinity of the original annotations.
In the second stage, we randomly sample potential \classname{marine debris} pixels, as well, as markers for non-debris pixels (class \classname{other}) in the remaining parts of the image.
These markers are the starting points of a random walk segmentation algorithm \citep{grady2006random}, which is a fast algorithm that requires a few labeled pixels as markers. The markers are assumed to be accurately annotated, while the pixels between the markers are uncertain and are then annotated by an underlying anisotropic diffusion process that ensures that homogeneous areas are assigned to the same class. Crucially, one set of parameters (homogeneity criterion, buffer size, marker sampling frequency) of the random walker algorithm leads to one potential debris map. Therefore, we vary those parameters and average all maps to capture the underlying undefinedness of the borders of marine debris, as shown in the bottom row of \cref{fig:label_refinement_module}.

\subsubsection{MARIDA}

The Marine Debris Archive (MARIDA) was collected by \citet{kikaki2022marida} for developing and evaluating machine learning algorithms for marine debris detection. MARIDA contains 63 temporally overlapping {Sentinel-2} scenes from 12 distinct regions. In total, \num{6672} polygons were annotated, of which \num{1882} are \classname{marine debris} and \num{2447} \classname{marine water}. The remaining \num{2343} polygons are annotated in one of 13 further classes with between 24 and 356 annotations each that we do not use in this study.
%
We use MARIDA as additional training, validation, and evaluation data source, but consider only patches annotated as \classname{marine debris} (positive class) and treat instances of \classname{marine water} as negatives.
The MARIDA dataset contains {Sentinel-2} imagery with 11-bands that have been atmospherically corrected with the ACOLITE \citep{vanhellemont2018atmospheric} algorithm. In this work, we want to apply our detector on 12-band {Sentinel-2} imagery that had been atmospherically corrected with Sen2COR \citep{main2017sen2cor}, as is readily available, for instance, in Google Earth Engine \citep{gorelick2017google}. This avoids reprocessing additional imagery after download and simplifies the application on new scenes. To harmonize this dataset, we re-downloaded all {Sentinel-2} scenes from Google Earth Engine to retrieve 12-band imagery for MARIDA compatible with the other datasets. Like FloatingObjects, we use the atmospherically corrected L2A {Sentinel-2} imagery whenever available. 
We also excluded one scene near Durban from MARIDA (named \regionname{S2 24-4-19 36JUN}) to avoid spatial overlap, and potential positive biases with our evaluation scene described later in \cref{sec:evaluation}.

\subsubsection{S2Ships}
\label{sec:data:s2ships}

Ships and their wakes can cause false positive predictions of marine debris, as reported by \citet{mifdaletal2020towards}. We decided to explicitly add images of ships without any annotated marine debris as negative examples. We use the S2Ships dataset of \citet{ciocarlan2021ship}, which segmented ships with {Sentinel-2} imagery.
In our training pipeline, we retrieve these ship positions, load an image centered on each ship and show it to our detector during training with a negative prediction mask indicating the class \classname{other}.

\subsection{Validation and Evaluation Sites}
\label{sec:evaluation}

For finding the best neural network design and hyperparameters (i.e., validation), as well, as for the final independent evaluation, we used datasets with high-quality annotations. 
For both sets, we combine the MARIDA datasets, according to their validation and evaluation partitioning schemes, with a refined version of the FloatingObjects dataset that we describe in the next \cref{sec:refinedFlobs}. 
For further qualitative evaluation, we additionally use imagery from the Plastic Litter Projects 2021 and 2022, detailed further in \cref{sec:plp}. For both validation and evaluation datasets, we focus on using accurate annotations and we select only sites with a high probability of plastic pollution specifically for final evaluation, as detailed further in the next sections.

\subsubsection{RefinedFloatingObjects}
\label{sec:refinedFlobs}

We create a refined version of the FloatingObjects dataset (\cref{sec:flobs}) with less label noise, by re-annotating some a subset of FloatingObjects regions by individual point locations of which we are certain that they are localized accurately on visible marine debris in the imagery. We conduct this annotation in Google Earth Engine (GEE) \citep{gorelick2017google} and select the subset of regions named \regionname{lagos 20190101}, \regionname{neworleans 20200202}, \regionname{venice 20180630}, \regionname{accra 20181031}. We also included two new areas, which are \regionname{marmara 20210519} and \regionname{durban 20190424}. By carefully annotating these areas, we are confident that we captured the precise location of the class \classname{marine debris} in these {Sentinel-2} scenes. To train a model, we also need examples for the negative \classname{other} class to calculate accuracy scores that capture a diverse set of negatives, like open water, land, coastline, and ships, that likely confuse the model. To obtain these negative examples, we iteratively added negative examples by monitoring the result of a smileCART \citep{breiman1984cart} classifier implemented online in Google Earth Engine. This classifier serves as a proxy antagonist to us as labelers, i.e., it will highlight areas that appear like \classname{marine debris} and will be checked by annotators.
We explicitly added new negative examples in locations where this proxy classifier incorrectly predicted marine debris.
Hence, we captured meaningful negative point locations of the \classname{other} class that was difficult to distinguish from the annotated \classname{marine debris} by the smileCart classifier.

At validation and evaluation time, we extract a $\SI{128}{\pixel} \times \SI{128}{\pixel}$ patches centered on each of these annotated points that are labeled as either \classname{marine debris} (positive) or \classname{other} (negative). We can only be certain about the class at the precise annotations of the point in the center of each image patch. Hence, we first segment the entire patch using the semantic segmentation model but then extract the prediction only at the center pixel corresponding to the annotated point for accuracy estimation. This selection effectively simplifies the segmentation problem to a classification problem at the center of the image patch. It allows us to use standard classification metrics to measure the accuracy (described in \cref{sec:metrics}). 

Among the six regions in this dataset, we use the {Sentinel-2} scenes \regionname{lagos 20190101}, \regionname{neworleans 20200202}, \regionname{venice 20180630}, \regionname{marmara 20210519} for validation, as we are not certain about the composition of the visible marine debris in these images. For instance, \regionname{marmara 20210519} likely contains floating algae (sea snot), as it coincides with reported algae blooms \citep{dwmarmara}\new{ which are often present in this area \citep{hu2022spectral}.} \new{We use the accurate annotations of this generic marine debris in these areas to calibrate the model hyperparameters, such as the classification threshold, before final evaluation.}

\begin{figure}
    \centering
    \includegraphics[width=\textwidth]{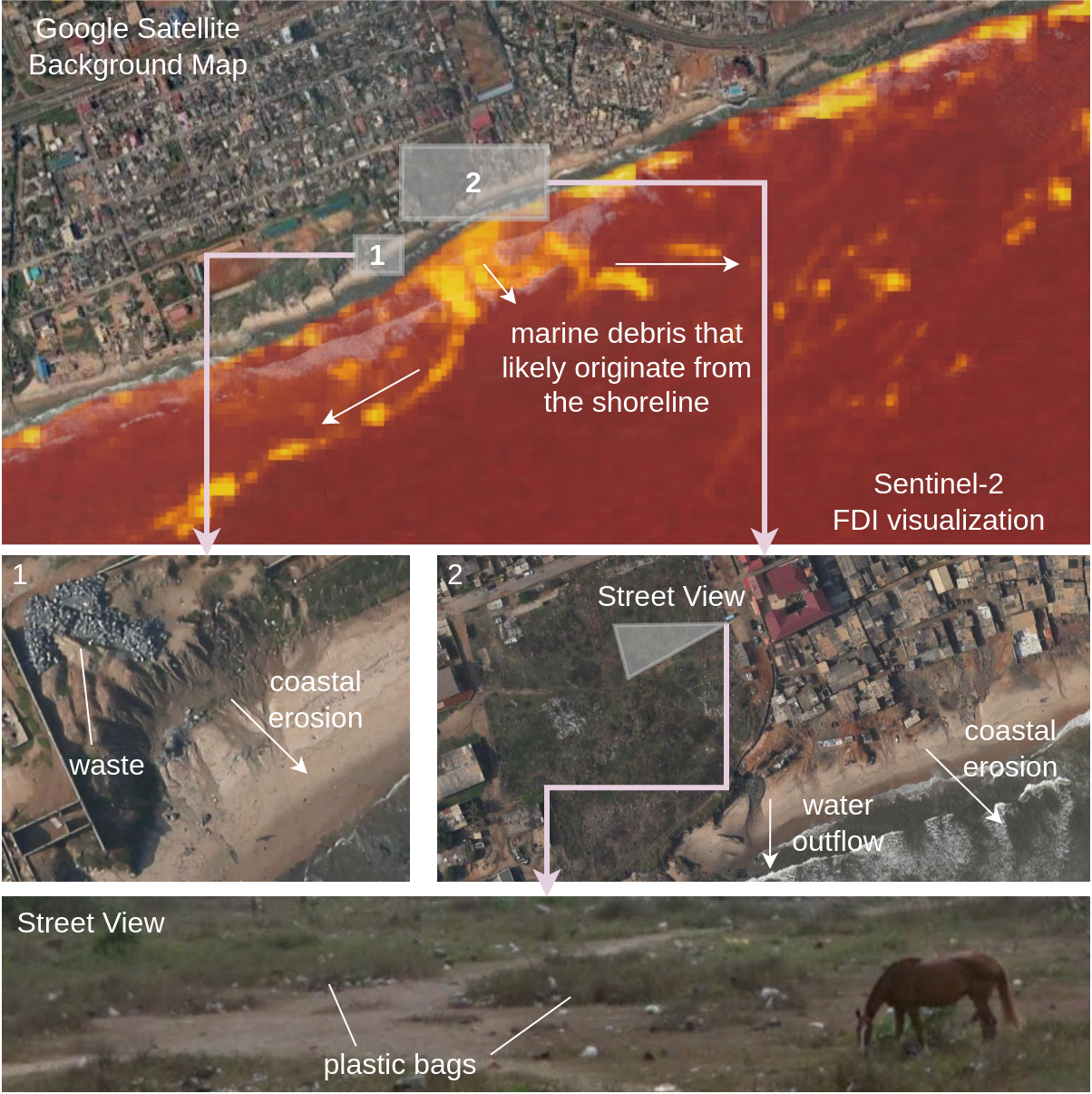}
    \caption{Evaluation scene in Accra, Ghana. The top row shows an FDI visualization of the {Sentinel-2} image of October \nth{31} 2018, where marine debris is washed into the open waters. Closer investigations with high-resolution satellite images (center row) show that coastal erosion is present, and this area is generally polluted with human litter. This is also confirmed by a Google Street View image shown on the bottom row.}
    \label{fig:accra}
\end{figure}

\begin{figure}
	\centering
	\begin{subfigure}{.355\textwidth}
		\includegraphics[width=\textwidth]{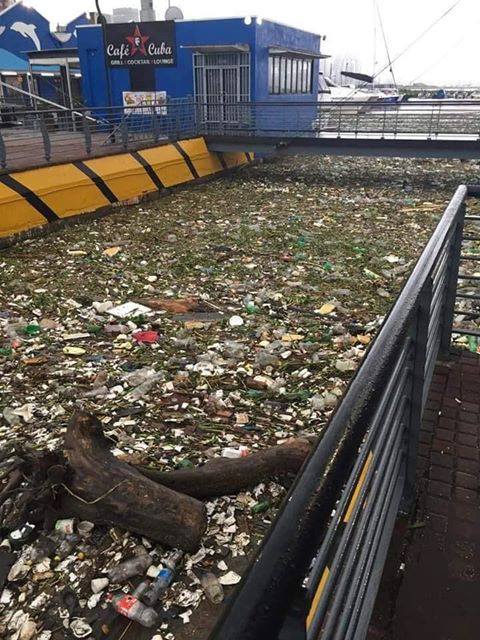}
		\caption{Photo: Ash Erasmus}
	\end{subfigure}
        \hfill
	\begin{subfigure}{.635\textwidth}
		\includegraphics[width=\textwidth]{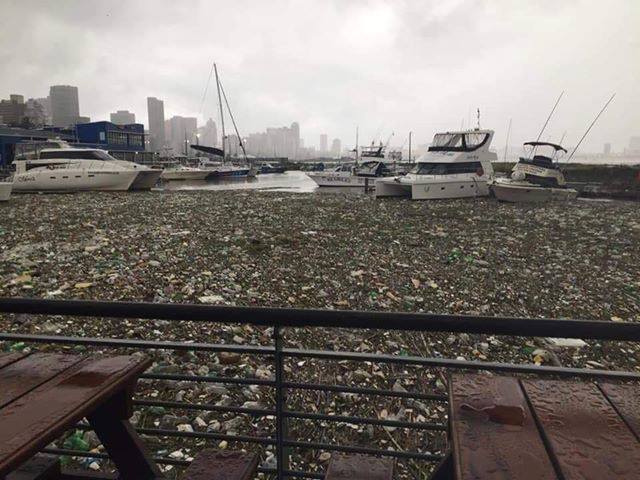} 
		\caption{Photo: Ash Erasmus} 
	\end{subfigure}
	\begin{subfigure}{\textwidth}
		\includegraphics[width=\textwidth]{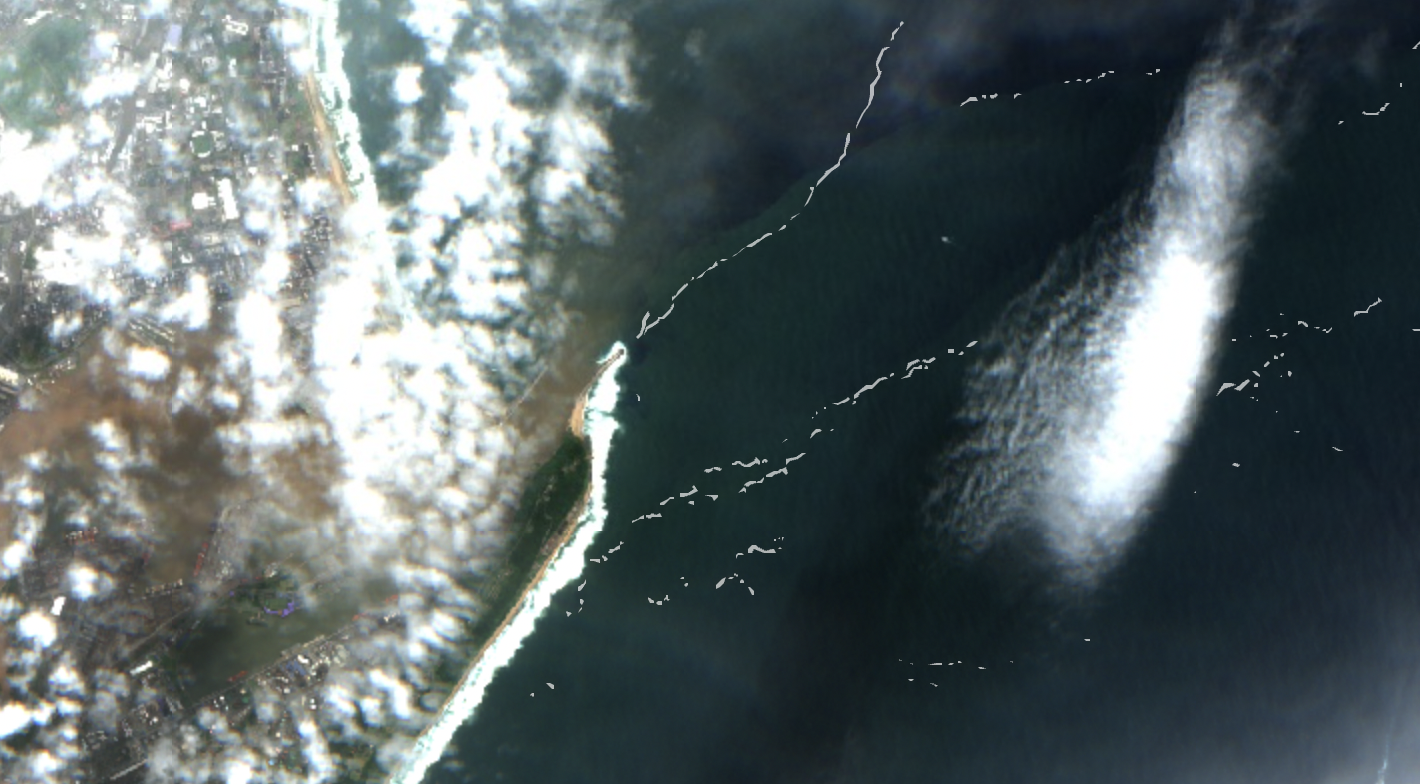}
		\caption{Sentinel-2 evaluation scene with debris annotations.}
		\label{fig:durban:s2}
	\end{subfigure}
	
	\caption{Evaluation scene from Durban, South Africa. Additional imagery shared by local news and social media (top row) show the level of plastic pollution on \nth{24} of April 2019. The {Sentinel-2} image (bottom image) shows the corresponding {Sentinel-2} scene we use for evaluation.}
	\label{fig:durban}
\end{figure}

For evaluation, we use the scenes \regionname{accra 20181031} and \regionname{durban 2019042}, as these areas very likely contain plastics in the marine debris:
\begin{itemize}
    \item \textbf{Evaluation Scene Accra, Ghana, 2018-10-31. }
Beach surveys in 2013 showed that plastic materials made up the majority of 63.72\% of marine debris washed onto evaluated beaches \citep{van2016empirical}. A recent study \citep{Pinto2023} estimated the daily plastic mass transport of plastic in the Odaw river running through Accra into the sea between \num{140} and \num{380} kilogram per day. 
Qualitatively, one particular area in this {Sentinel-2} scene, shown in  \cref{fig:accra} (top), shows an outwash of debris from the coast. In this image, the marine debris are visible in yellow (high floating debris index FDI). We show a high-resolution background map from Google Satellites for land and shoreline to provide a reference. Two zoomed-in areas (named 1 and 2 in \cref{fig:accra}) show that coastal erosion is visible alongside waste and sewage outflows aggregations. Finally, a Google Street View image (bottom row of \cref{fig:accra}) further confirms this area's general pollution level. Only a Sentinel-2 image at the top-of-atmosphere processing level (L1C) is available in Google Earth Engine in Accra.
\item \textbf{Evaluation Scene Durban, South Africa, 2019-04-24}. 
This evaluation scene was first identified by \citet{biermann2020finding}, who used social media and news reports to select areas of plastic pollution.
It covers marine debris that likely contains plastic litter from a flood event in Durban following heavy rainfall starting on April \nth{18} 2019.
This flood discharged large quantities of debris into the harbor of the Durban Metropole, as shown in \cref{fig:durban}.
We acquired one {Sentinel-2} image from April 24th, shown in \cref{fig:durban:s2}, where visible debris originates from the harbor area (highlighted in gray).
The debris in this image likely contains plastic litter.
This image is particularly difficult to predict, as clouds and haze from former precipitations are still visible in this scene. The patches of marine debris visible in the FDI representation are less pronounced than in the Accra scene, which has more clearly identifiable objects. In this area both top-of-atmosphere (L1C) and bottom-of-atmosphere (L2A) Sentinel-2 images are available. We compare the model performance on both versions later in \cref{sec:confusions}.

\end{itemize}

\subsubsection{Plastic Litter Projects}
\label{sec:plp}

The third evaluation area covers {Sentinel-2} data showing explicitly deployed debris targets in the Plastic Litter Projects of 2021 and 2022 \citep{topouzelis2019detection,plp2019,PLP2021} on the island of Lesbos, Greece.
In 2021, one \SI{28}{\meter} diameter high-density polyethylene (HDPE) mesh was deployed on June \nth{8} 2021, followed by a \SI{28}{\meter} wooden target on June \nth{17} 2021. Both were visible during 22 {Sentinel-2} satellite overpasses until \nth{7} of October 2021.
In the Plastic Litter Project 2022, one \SI{5}{\meter} $\times$ \SI{5}{\meter} inflatable PVC target, alongside two \SI{7}{\meter} diameter HDPE meshes were deployed on June \nth{16} 2022. One HDPE mesh was cleaned regularly, while the other was subject to natural fouling and algae. The objects were deployed until the \nth{11} of October 2022 and were visible in 23 {Sentinel-2} acquisitions. Additional smaller \SI{1}{\meter\squared} and  \SI{3}{\meter\squared} targets were also deployed throughout the project phase to study visibility and the material's decomposition in water but were too small to be visible in the {Sentinel-2} scenes.
We use the {Sentinel-2} data of the 2021 campaign to qualitatively test the ability of our detector and comparison models to detect the deployed targets in the {Sentinel-2} imagery.

\subsection{Marine Debris Detector Implementation}
\label{sec:model}

This section describes the implementation of the Marine Debris Detector as a deep segmentation model that inputs a 12-channel {Sentinel-2} image and estimates the probability of marine debris's presence for each pixel. 

\begin{figure}
	\centering
	\includegraphics[width=\textwidth]{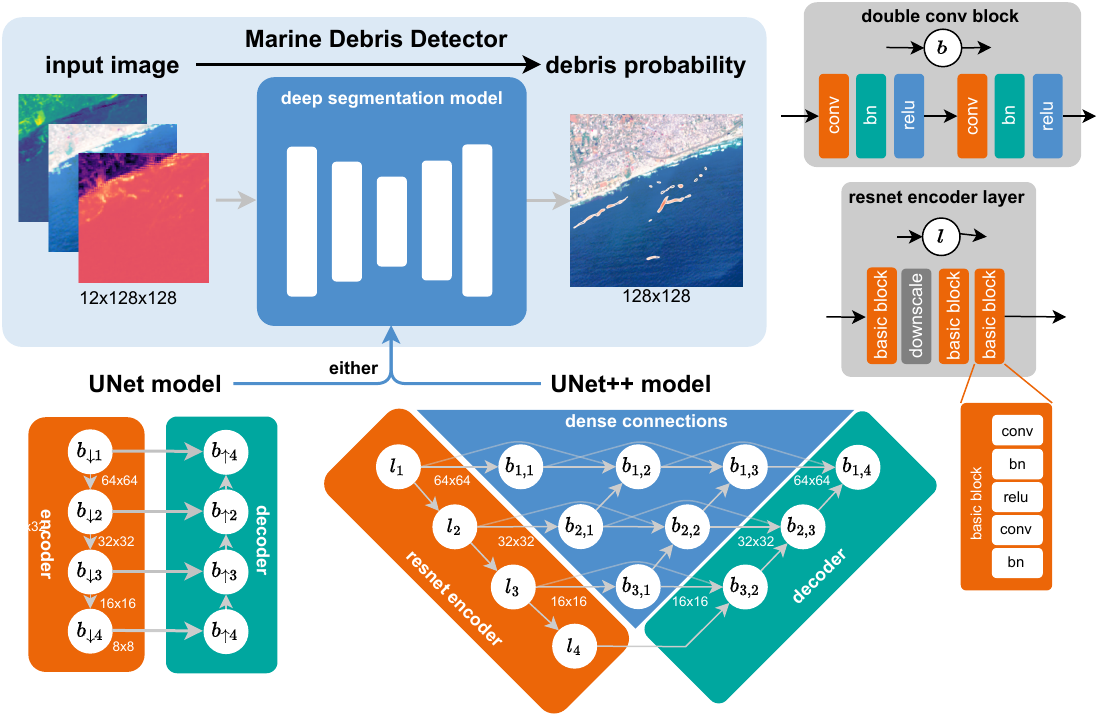}
	\caption{Schematic of the 
Marine Debris Detector implementation with an underlying \modelname{Unet} \citep{ronneberger2015u} or \modelname{Unet++} \citep{zhou2018Unetpp} segmentation model. A 12-channel input image (top-left) is encoded to hidden feature representations in several levels of resolution (vertical pathways) and decoded to a probability of marine debris (top-right). Higher-resolution pathways ensure that the resulting segmentation map is fine-grained, while lower-resolution encode global information on the entire scene. \modelname{Unet++} \citep{zhou2018Unetpp} extends the original \modelname{unet} \citep{ronneberger2015u} by adding additional dense connections in the skip pathways indicated in blue.}
	\label{fig:Unetpp}
\end{figure}

\subsubsection{Segmentation Model Architectures}

We implemented the \modelname{UNet} \citep{ronneberger2015u} and \modelname{Unet++} \citep{zhou2018Unetpp} architectures, as shown in \cref{fig:Unetpp}.
The \modelname{Unet} segmentation model of \citet{ronneberger2015u} was developed for medical image segmentation and is heavily used in remote sensing due to the fine-grained segmentation masks it can produce. The success of the \modelname{Unet} is strongly related to its early skip connections, which help maintain the details of the image in the final map. As such, skip connections enable the propagation of a high-resolution representation of the input image through the entire network. This network was the one used previously by \citet{mifdaletal2020towards} for marine debris detection.

The \modelname{Unet++}  \citep{zhou2018Unetpp} variant extends the original \modelname{Unet} by replacing the original encoder with a \modelname{ResNet} \citep{he2016deep} with four blocks (indicated as $l_i$). 
\modelname{ResNets} are the de-facto standard feature extractor in computer vision, as they can learn complex representation while requiring fewer weights than many earlier networks. The decoder consists of three double-convolutional blocks (indicated with $b_i$). Each block consists of two convolution-batchnorm-relu transformations. While the original \modelname{Unet} directly connects the output of each encoder layer with the corresponding decoder layer of same resolution, the \modelname{Unet++} adds additional double-conv blocks in these skip pathways that are connected densely in the spirit of \modelname{DenseNet} neural networks \citep{zhu2017densenet}. 

\subsubsection{Implementation and Training Details}

We train \modelname{Unet} and \modelname{Unet++} models with a learning rate of \num{0.01} and weight decay \num{1e-6} for 100 epochs.
The \modelname{Unet} implementation in this work has \num{31} million trainable parameters, while the \modelname{Unet++} has \num{26} million parameters.
Regarding the label refinement module (\cref{sec:flobs}), we compute multiple refined segmentation masks with different parameters and choose a buffer size of 0, 1, or 2 pixels, the $\beta$-parameter of the random walker (a penalization coefficient for the walker motion) of 1 or 10, and the {marker} density for marine debris of 5\%, 25\%, 50\% or 75\% (the density of \classname{other} {markers} is fixed at 5\%). Combined with the original mask, this yields 25 different target masks consistent with the hand annotations and the FDI image but of varying shapes and sizes, as shown in the bottom row of \cref{fig:label_refinement_module}. During training, we choose one of these target masks randomly, which, in our opinion, reflects best the undefined borders of the marine debris that we aim to detect and acts as a form of natural label-data augmentation.
During training, we monitor the area under the ROC curve (AUROC) on the refined FloatingObjects dataset (\cref{sec:refinedFlobs}) and MARIDA validation set. We store the model weights each time the highest (best) validation AUROC has been reached.
We observe that the model systematically underestimates the probability of marine debris due to a heavy class imbalance in the training data. This results in a low precision but high recall when we assign the class \classname{marine debris} for probability scores above 0.5.
We counteract this imbalance by calibrating the classification threshold to balance precision and recall on the validation set.

For the \modelname{Unet++} model, we trained models from different random seeds with validation-optimal thresholds of \num{0.132} \num{0.0639}, and \num{0.0254} during the experiments shown in this paper.  For the \modelname{Unet}, the thresholds were \num{0.0895}, \num{0.0712}, and \num{0.0643}.

Training a \modelname{Unet++} and \modelname{Unet} took eight and nine hours on an NVIDIA RTX 3090 graphics card with multi-threaded data loading with 32 workers. The estimated carbon footprint for one model training run was $2.8$ kg.eCO$^2$.


\subsection{Comparison Methods}
\label{sec:comparisonmodels}

We compare models trained within our training framework to approaches from recent literature. 
In particular, the \modelname{Unet} trained by \cite{mifdaletal2020towards} on the original FloatingObjects dataset, and a Random Forest classifier, denoted by \modelname{rf}, trained on the original MARIDA dataset \citep{kikaki2022marida}.
For the \modelname{Unet}, we use the provided pre-trained weights for their model.
Similarly to our segmentation models, we also determine the best classification threshold based on the validation set to achieve results with balanced precision and recall, which is \num{0.039}.
For the random forest classifier (\modelname{rf}), we train the random forest on 11 Sentinel-2 bands, as in the original paper with 12 output classes, and combine the predictions into a binary scheme by considering \classname{marine debris} as the positive class and treat all other 11 non-debris classes as \classname{other}.
In the results section, we denote these two models as \modelname{Unet} and \modelname{rf} and indicate that they have been trained on the \enquote{original data} of their respective papers.

We also train the random forest on the combined training dataset described in \cref{sec:trainingdata}, which we denote as \enquote{trained on our dataset}.
For the random forest, we use an identical feature extraction pipeline as described in \citet{kikaki2022marida}, which results in 26 features containing the original spectral bands, spectral indices, and textural features.
As the random forest is a pixel-wise classifier, we treat each pixel separately and create a roughly balanced training pixel dataset set from our image training dataset. We select five positive pixels (annotated as \classname{marine debris}) and five negative \classname{other} pixels from each image.
This results in a \num{70000} training pixels. As for to the other comparison approaches, we tune the classification threshold based on the validation dataset, which is \num{0.663}.

\subsection{Evaluation Metrics}
\label{sec:metrics}

We compare all models trained on \enquote{original data} and \enquote{our dataset} on several metrics on the evaluation sets of Durban, Accra, and the MARIDA test partitions.
\begin{itemize}
    \item We include the overall \metric{accuracy} ratio of correct classifications to total samples. It is straightforward to interpret, but susceptible to class imbalance. Our selected validation and evaluation sets, however, have a general balance between positive and negative samples.
	\item \metric{f-score} is the harmonic mean between precision and recall that, in contrast to individual precision and recall scores, is more robust to the choice of the classification threshold.
	\item The area under the receiver operator curve (\metric{auroc}) is a metric that is independent of the classification thresholds but easily saturates for relatively accurate classifiers with values close to 1.
	\item The \metric{jaccard} index, also known as intersection over union, is commonly used for object detection and measures the number of intersections of two sets (predictions and ground truth) divided by their union.
	\item The \metric{kappa} statistic compares two classifiers: the model and a randomly guessing baseline. Values of zero indicate that the tested model is not better than a random baseline, while positive correlations indicate that the tested model outperforms the trivial baseline.
\end{itemize}
Higher values are better for all metrics, and values of 1 indicate a perfect score.

\section{Results}
\label{sec:results}

We first compare the models quantitatively and qualitatively in \cref{sec:comparisons}.
We then predict one entire {Sentinel-2} scene (Durban) in \cref{sec:confusions} and quantify the false positive predictions on both bottom-of-atmosphere and top-of-atmosphere {Sentinel-2} imagery. 
In the final experiment \cref{sec:transferability}, we test how a re-trained 4-channel detector can predict marine debris on higher-resolution PlanetScope imagery, which can complement {Sentinel-2} imagery in practice.

\subsection{Numerical Comparisons}
\label{sec:comparisons}

{
	\onehalfspacing
\begin{table}[]
    \footnotesize\sffamily
    \centering
    \begin{subtable}{\textwidth}
        \centering
        \begin{tabular}{lrlrcll}
        
        \toprule
            Accra \\
            trained on & \multicolumn{2}{l}{original data} & \multicolumn{4}{c}{our train set} \\
                        & \modelname{rf} & \modelname{unet} & \modelname{rf} & \modelname{Unet} & \modelname{Unet++} & \modelname{Unet++} no-ref\\
                        \cmidrule(lr){2-3}\cmidrule(lr){4-7}
            \metric{accuracy} & 0.653 & 0.882 & 0.680 & {0.924} $\pm$ 0.016 & {0.930} $\pm$ 0.016 & \textbf{0.948} $\pm$ 0.008 \\
            \metric{f-score}  & 0.464 & 0.871 & 0.545 & {0.920} $\pm$ 0.018 & {0.926} $\pm$ 0.018 & \textbf{0.948} $\pm$ 0.008 \\
            \metric{auroc}    & 0.246 & 0.965 & 0.899 & {0.978} $\pm$ 0.008 & {0.981} $\pm$ 0.006 & \textbf{0.989} $\pm$ 0.005 \\
            \metric{jaccard}  & 0.302 & 0.772 & 0.374 & {0.852} $\pm$ 0.030 & {0.862} $\pm$ 0.031 & \textbf{0.900} $\pm$ 0.014 \\
            \metric{kappa}    & 0.301 & 0.764 & 0.357 & {0.848} $\pm$ 0.031 & {0.859} $\pm$ 0.031 & \textbf{0.897} $\pm$ 0.017 \\
    
            \bottomrule
        \end{tabular}
    \end{subtable}

        \vspace{.5em}
    \begin{subtable}{\textwidth}
    \centering
    
        \begin{tabular}{lrlrcll}
        \toprule
            Durban  \\
            trained on & \multicolumn{2}{l}{original data} & \multicolumn{4}{c}{our train set} \\
                        & \modelname{rf} & \modelname{Unet} & \modelname{rf} & \modelname{Unet} & \modelname{Unet++} & \modelname{Unet++} no-ref  \\
                        \cmidrule(lr){2-3}\cmidrule(lr){4-7}
            \metric{accuracy} & 0.781 & 0.587 & 0.811 &  0.908 $\pm$ 0.010 &   \textbf{0.934} $\pm$ 0.018 & 0.905 $\pm$ 0.011 \\
            \metric{f-score}  & 0.105 & 0.497 & 0.708 &  0.756 $\pm$ 0.032 &   \textbf{0.837} $\pm$ 0.053 & 0.776 $\pm$ 0.026 \\
            \metric{auroc}    & 0.376 & 0.765 & 0.862 &  0.850 $\pm$ 0.030 &   \textbf{0.914} $\pm$ 0.018 & 0.886 $\pm$ 0.053 \\
            \metric{jaccard}  & 0.055 & 0.330 & 0.548 &  0.609 $\pm$ 0.042 &   \textbf{0.722} $\pm$ 0.048 & 0.635 $\pm$ 0.034 \\
            \metric{kappa}    & 0.082 & 0.245 & 0.569 &  0.704 $\pm$ 0.037 &   \textbf{0.797} $\pm$ 0.063 & 0.717 $\pm$ 0.031 \\
    
            \bottomrule
        \end{tabular}

    \end{subtable}

    \vspace{.5em}
    \begin{subtable}{\textwidth}
\centering

        \begin{tabular}{lrlrcll}
        \toprule
            \multicolumn{4}{l}{Marida-test set}  \\ 
            trained on & \multicolumn{2}{l}{original data} & \multicolumn{4}{c}{our train set} \\
                        & \modelname{rf} & \modelname{Unet} & \modelname{rf} & \modelname{Unet} & \modelname{Unet++} & \modelname{Unet++} no-ref  \\
                        \cmidrule(lr){2-3}\cmidrule(lr){4-7}
            \metric{accuracy} & 0.697 & 0.838 & 0.811 & \textbf{0.865} $\pm$ 0.006 & \textbf{0.867} $\pm$ 0.005 & 0.851 $\pm$ 0.006 \\
            \metric{f-score}  & 0.288 & 0.701 & 0.708 & \textbf{0.741} $\pm$ 0.012 & \textbf{0.749} $\pm$ 0.009 & 0.710 $\pm$ 0.015 \\
            \metric{auroc}    & 0.488 & 0.764 & 0.862 & \textbf{0.738} $\pm$ 0.012 & \textbf{0.746} $\pm$ 0.021 & 0.733 $\pm$ 0.006 \\
            \metric{jaccard}  & 0.168 & 0.539 & 0.548 & \textbf{0.589} $\pm$ 0.015 & \textbf{0.598} $\pm$ 0.012 & 0.551 $\pm$ 0.018 \\
            \metric{kappa}    & 0.197 & 0.593 & 0.569 & \textbf{0.654} $\pm$ 0.016 & \textbf{0.661} $\pm$ 0.012 & 0.615 $\pm$ 0.017 \\
    
            \bottomrule
        \end{tabular}
    
    	\begin{subfigure}{\textwidth}
    		\centering
    		\includegraphics[width=.75\textwidth]{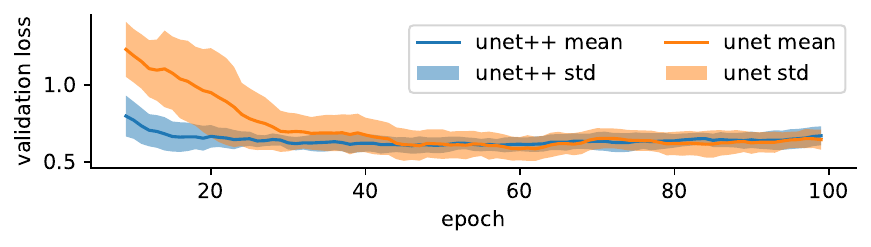}
    	\end{subfigure}

    \end{subtable}

        \caption{Quantitative comparison of models trained on original data (\modelname{rf} \citep{kikaki2022marida}, \modelname{Unet} \citep{mifdaletal2020towards}), versus models trained on the training data compiled in this work. We also test a \modelname{Unet++} model without label refinement module, indicated by the \enquote{no-ref} suffix in the last column. The bottom plot shows the validation loss during training of three \modelname{Unet++} and \modelname{Unet} models, each. The \modelname{Unet++} finds an optimum earlier and has less variance (shown in $1\sigma$ standard deviation) between the models in the early states of training.} 
    \label{tab:quantitativecomparison}
    
\end{table}
}

\cref{tab:quantitativecomparison} shows the quantitative results of \modelname{rf} and \modelname{Unet} models trained on the respective original data in comparison to \modelname{rf}, \modelname{unet}, and \modelname{unet++} trained with our training setting on the combined training dataset and refinement strategies described in \cref{sec:trainingdata}.
We see that models trained in our combined training framework achieve the best accuracy metrics in all experiments including those where the label refinement is not used (column \enquote{no-ref}). 
As expected, the deep learning-based \modelname{UNet} and the \modelname{Unet++} models outperform the pixel-wise random forest classifier. This is likely due to the advantage of convolutional neural networks to learn spatial patterns within their convolutional perceptive field.
Both \modelname{Unet} and \modelname{Unet++} achieve equal accuracies within one standard deviation on the Marida test set, while the \modelname{Unet++} achieves a better accuracy on the Durban and Accra scenes. The label refinement module also improves the \modelname{Unet++} performance on Marida-test and Durban. However, on Accra, the best scores are achieved with a \modelname{Unet++} model without refinement module (indicated by \enquote{no-ref}).
For the remaining paper, we use the \modelname{Unet++} model in the Marine Debris Detector, as it has fewer parameters and finds an optimum earlier and more consistently between random seeds ($1\sigma$ standard deviation shown) than the \modelname{Unet} in the training process, as shown in the bottom plot of \cref{tab:quantitativecomparison}. 

\begin{figure}[]
    \centering
    \sffamily
    \def\predimage{ypred}
    \def\width{2cm}
    
    \resizebox{\textwidth}{!}{
    \begin{tabular}{llllllll}
    & \multicolumn{2}{l}{input (12-bands)} & target & \multicolumn{4}{l}{model predictions} \\
    \cmidrule(lr){2-3}\cmidrule(lr){4-4}\cmidrule(lr){5-8}
    & & & & \multicolumn{3}{c}{our training data} & FlObs-only \\
    \cmidrule(lr){5-6} \cmidrule(lr){7-7} \cmidrule(lr){8-8}
    & RGB & FDI & label & \modelname{Unet++} & no-ref & \modelname{rf} & \modelname{Unet} \\

    \rotatebox[origin=lB]{90}{Accra-1} &
    \includegraphics[width=\width]{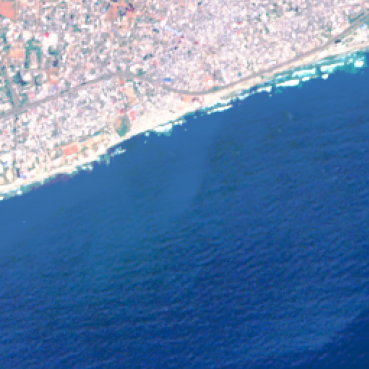} & 
	\includegraphics[width=\width]{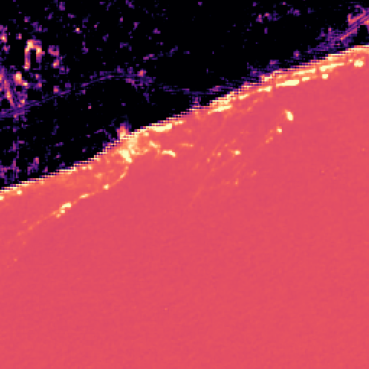} & 
	\includegraphics[width=\width]{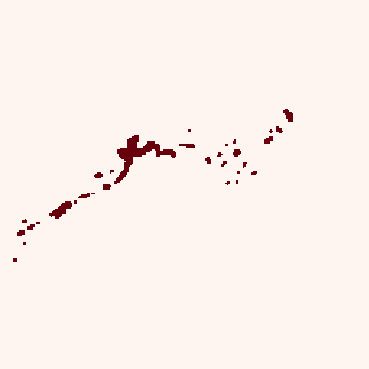} & 
	\includegraphics[width=\width]{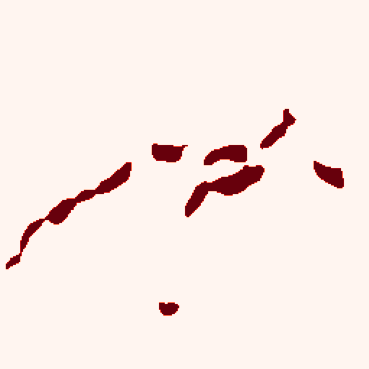} & 
	\includegraphics[width=\width]{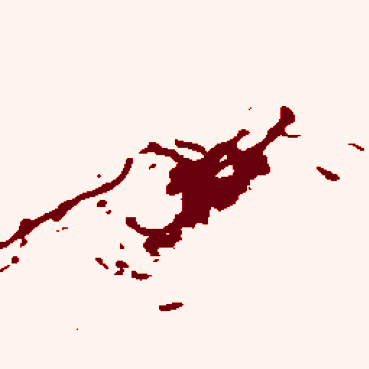} & 
	\includegraphics[width=\width]{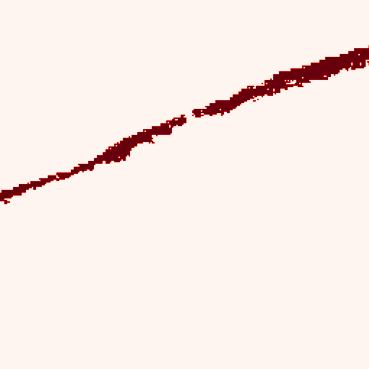} &
	\includegraphics[width=\width]{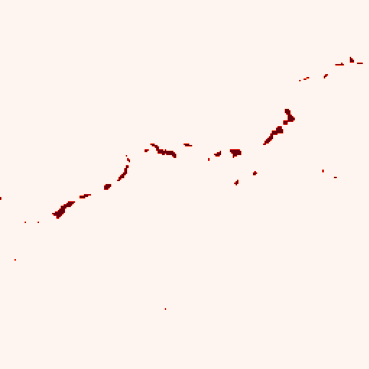} \\

    \rotatebox[origin=lB]{90}{Accra-2} &
    \includegraphics[width=\width]{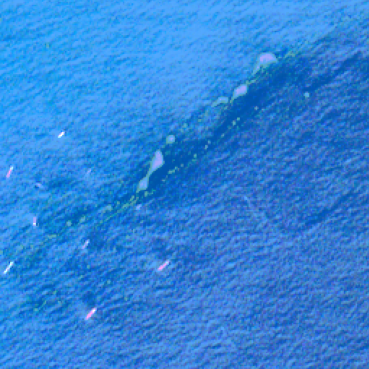} & 
	\includegraphics[width=\width]{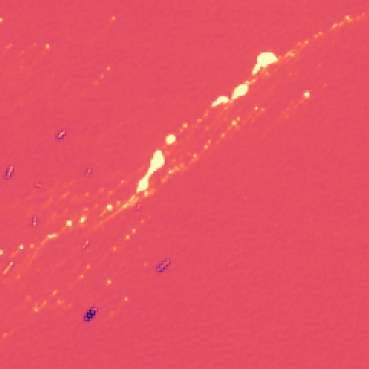} & 
	\includegraphics[width=\width]{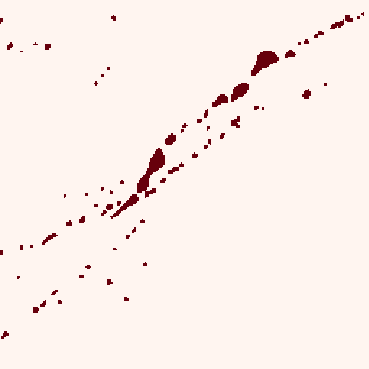} & 
	\includegraphics[width=\width]{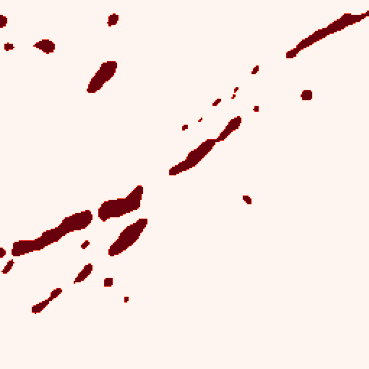} & 
	\includegraphics[width=\width]{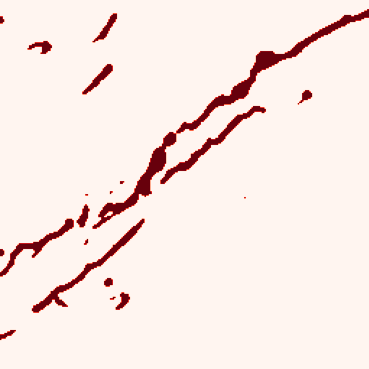} & 
	\includegraphics[width=\width]{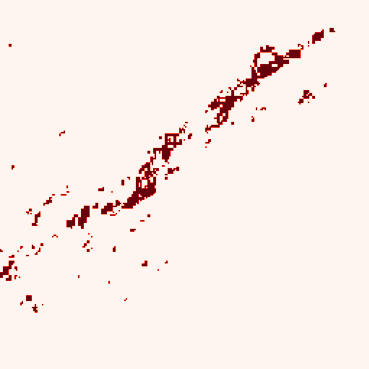} &
    \includegraphics[width=\width]{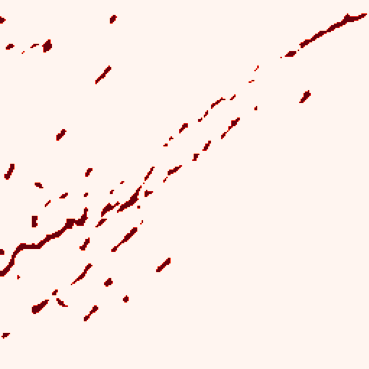} \\

    \rotatebox[origin=lB]{90}{Durban-1} &
    \includegraphics[width=\width]{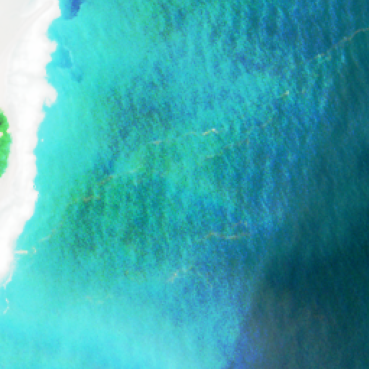} & 
	\includegraphics[width=\width]{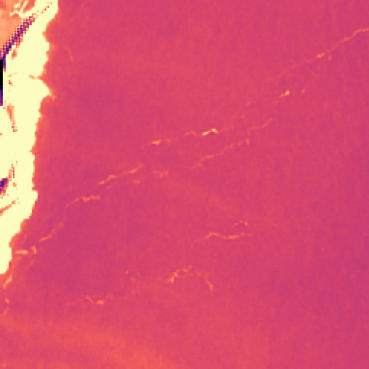} & 
	\includegraphics[width=\width]{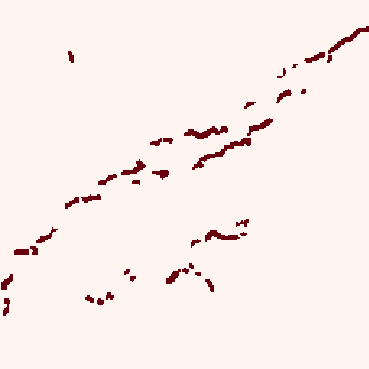} & 
	\includegraphics[width=\width]{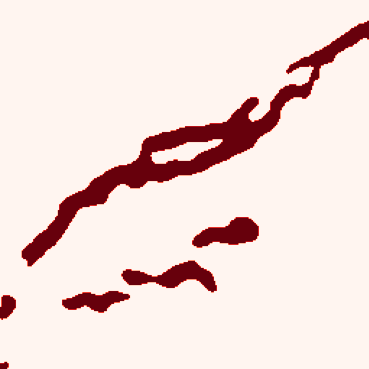} & 
	\includegraphics[width=\width]{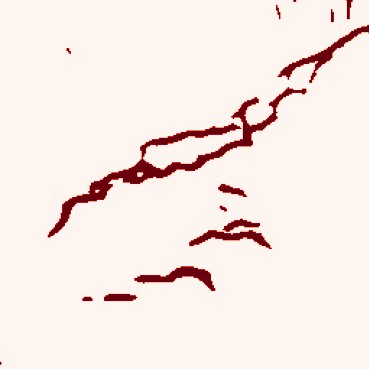} & 
    \includegraphics[width=\width]{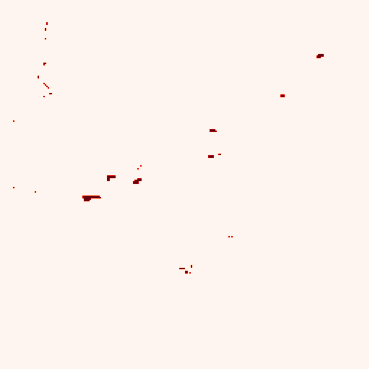} &
	\includegraphics[width=\width]{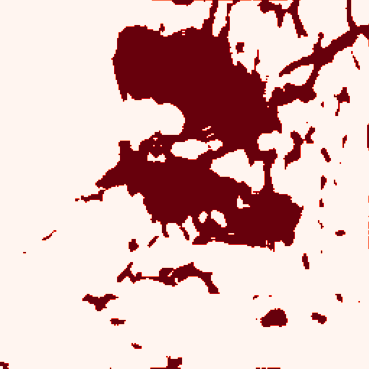} \\

    \rotatebox[origin=lB]{90}{Durban-2} &
    \includegraphics[width=\width]{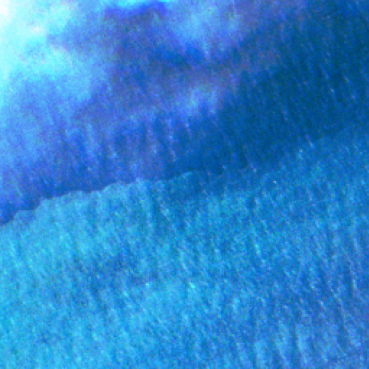} & 
	\includegraphics[width=\width]{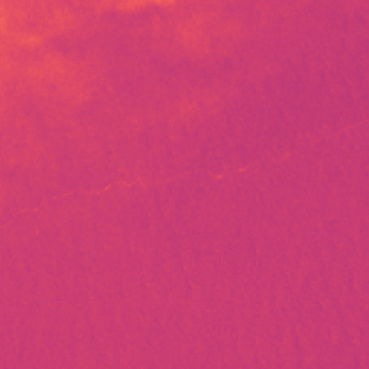} & 
	\includegraphics[width=\width]{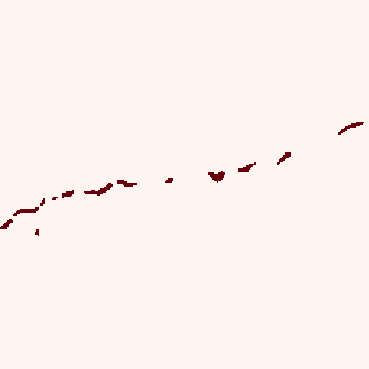} & 
	\includegraphics[width=\width]{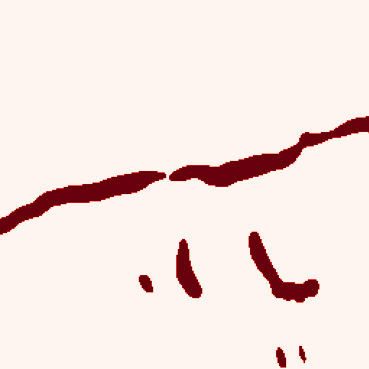} & 
	\includegraphics[width=\width]{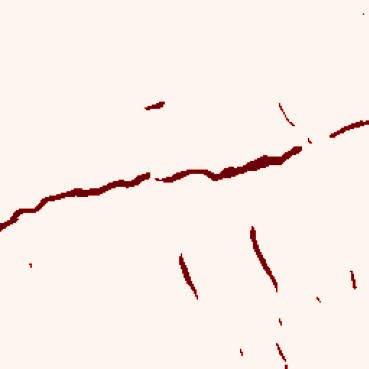} & 
 	\includegraphics[width=\width]{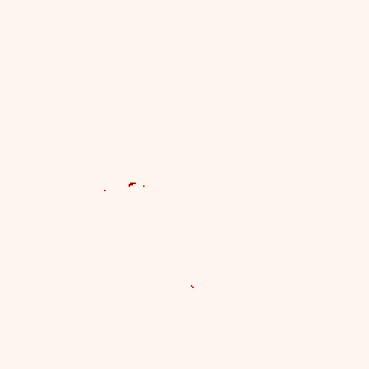} &
	\includegraphics[width=\width]{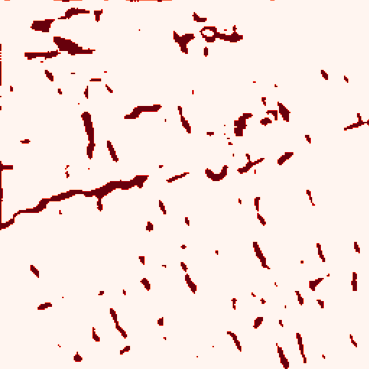} \\

    \rotatebox[origin=lB]{90}{Durban-3} &
    \includegraphics[width=\width]{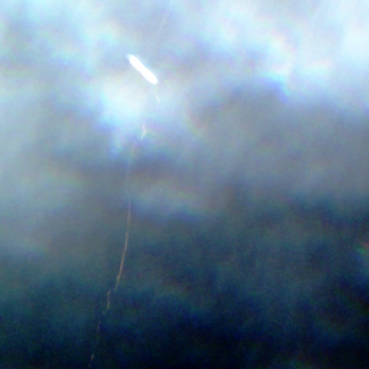} & 
	\includegraphics[width=\width]{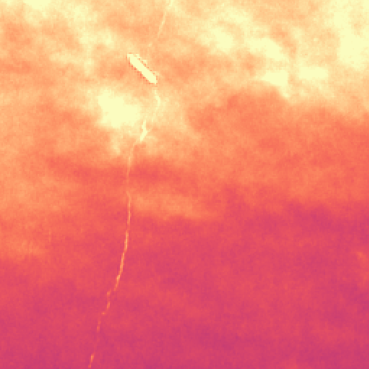} & 
	\includegraphics[width=\width]{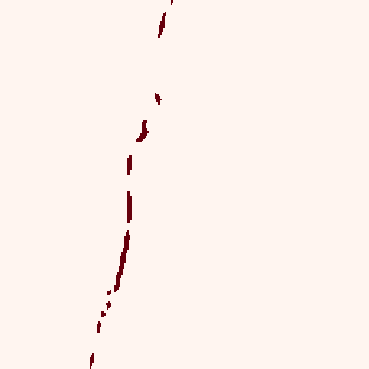} & 
	\includegraphics[width=\width]{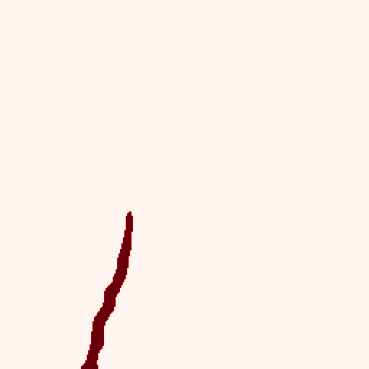} & 
	\includegraphics[width=\width]{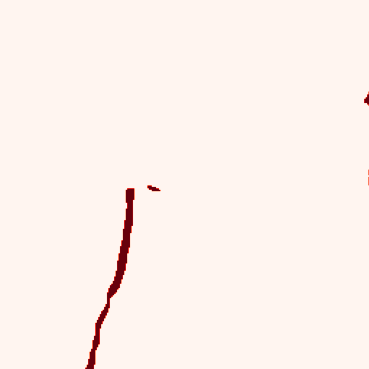} & 
    \includegraphics[width=\width]{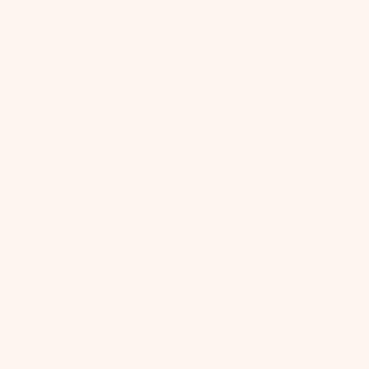} &
	\includegraphics[width=\width]{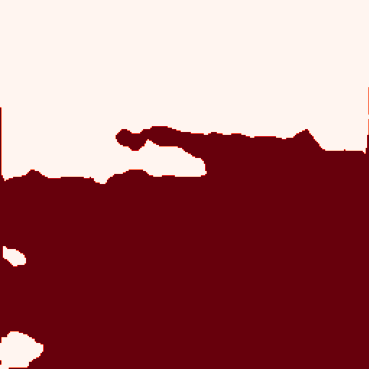} \\
    \end{tabular}
    }
    \caption{Qualitative predictions of the three models on images covering each \SI{2.56}{\kilo\meter} by \SI{2.56}{\kilo\meter} from the Accra and Durban sets. Our \modelname{Unet++} produces \classname{marine debris} predictions similar to the hand annotations (target/label) with the fewest false positives. An interactive qualitative comparison is available under  \url{https://marcrusswurm.users.earthengine.app/view/marinedebrisexplorer}}
    \label{tab:qualitativecomparison}
\end{figure}

\Cref{tab:qualitativecomparison} compares models qualitatively on selected \SI{256}{\pixel} $\times$ \SI{256}{\pixel} each patches covering \SI{2.56}{\kilo\meter} by \SI{2.56}{\kilo\meter}. The tiles are from the Accra and Durban evaluation scenes, where it is highly plausible that plastic pollution is present in marine debris.
We compare the \modelname{Unet++} model with and without label refinement, the random forest \modelname{rf} with features of \citep{kikaki2022marida}, trained on our dataset, and the \modelname{Unet} from \citet{mifdaletal2020towards} trained on the original FloatingObjects (FlObs) dataset only.
The first two columns show RGB and FDI representations of the multi-spectral {Sentinel-2} scenes. The third column shows hand-annotated masks (shown in red). 
We generally see the quantitative results mirrored in these qualitative examples, where the deep learning model trained on our combined training set produces the most truthful masks of floating marine debris.
While none of the models captured the hand annotations perfectly, the \modelname{Unet++} produced the visually most accurate predictions with the fewest false positives across most evaluation scenes. The \modelname{Unet++} without label refinement (indicated by \enquote{no-ref}) provides generally thinner predictions than the \modelname{Unet++} with refinement module, which we connect to the refinement module always enlarging the target mask of marine debris to some degree during training. 
In Accra-1, \modelname{Unet++} and \modelname{Unet} \citep{mifdaletal2020towards} capture the general location of the objects, while the random forest \modelname{rf} \citep{kikaki2022marida} detected natural waves along the entire coastline as marine debris. The \modelname{Unet++} without refinement module appears to merge multiple patches of debris here and does not accurately capture the individual objects.
Accra-2 shows several sargassum patches in between ships. Generally, all models predict these patches well, while still some ships are confused with marine debris.
The Durban scenes are more challenging and show more atmospheric perturbations through clouds and haze.
The \modelname{Unet++} predicts the general locations of the annotated marine debris well until the cloud coverage is too dense, as seen in Durban-3.
The original \modelname{Unet} \citep{mifdaletal2020towards} predicts a large number of false positives, which was also stated as a limitation in their original work. The random forest \modelname{rf} of \citet{kikaki2022marida} tends to under-predict the marine debris in all three Durban scenes and only identifies a few individual floating object patches in Durban-1.

\begin{figure}[]
    \centering
    \sffamily

    \def\width{1.2cm}
    \scriptsize
    
    	\centering
    	\def\kikakipath{figures/results/kikaki/plp2021}
    	\def\unetpponepath{figures/results/marinedebrisdetector/unet++1/plp2021}
    	\def\unetpptwopath{figures/results/marinedebrisdetector/unet++2/plp2021}
    	\def\unetppthreepath{figures/results/marinedebrisdetector/unet++3/plp2021}
    	\def\unetnorefonepath{figures/results/marinedebrisdetector/unet++1_no_label_refinement/plp2021}
    	\def\unetnoreftwopath{figures/results/marinedebrisdetector/unet++2_no_label_refinement/plp2021}
    	\def\unetnorefthreepath{figures/results/marinedebrisdetector/unet++3_no_label_refinement/plp2021}
    	\def\mifdalpath{figures/results/mifdal/plp2021}
    	\def\metric{yscore}
    	\begin{tabular}{crcccccccc}
    		
    		& RGB & FDI & \multicolumn{3}{c}{with label refinement} & \multicolumn{3}{c}{no label refinement} \\ 
    		&     & & seed 1 & seed 2 & seed 3 & seed 1 & seed 2 & seed 3 \\ 
    		 
    		\rotatebox[origin=lB]{90}{June \nth{11}}
    		& \includegraphics[width=\width]{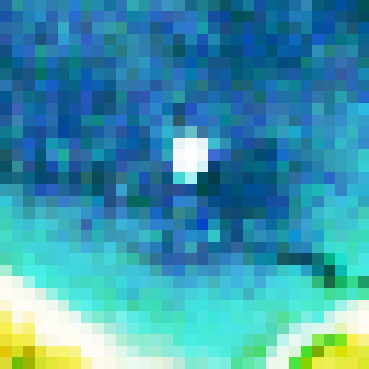} 
    		& \includegraphics[width=\width]{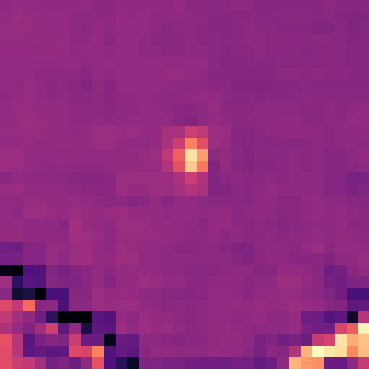} 
    		& \includegraphics[width=\width]{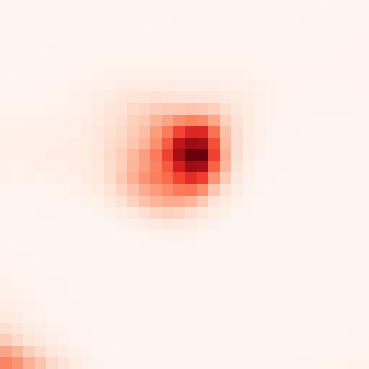} 
    		& \includegraphics[width=\width]{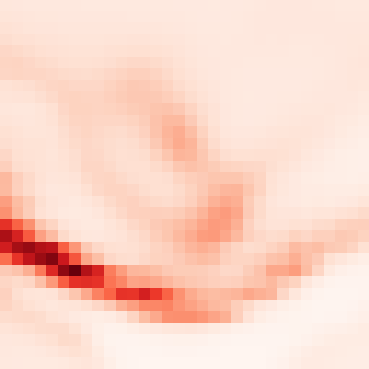} 
    		& \includegraphics[width=\width]{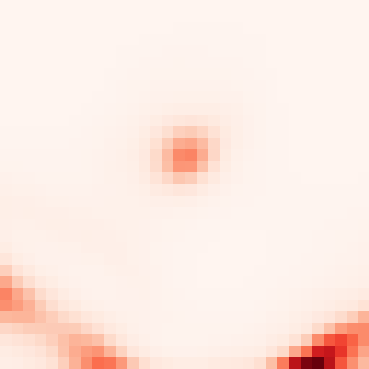} 
    		& \includegraphics[width=\width]{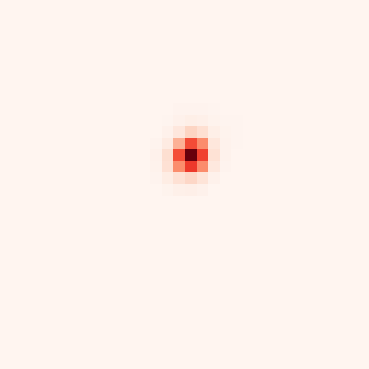} 
    		& \includegraphics[width=\width]{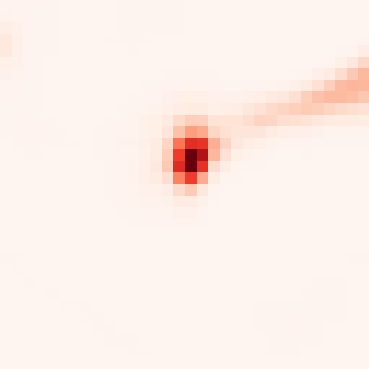} 
    		& \includegraphics[width=\width]{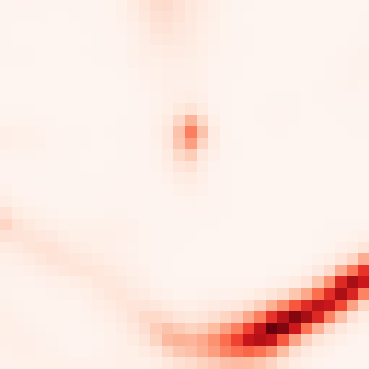} 
    		\\
    		
    		\rotatebox[origin=lB]{90}{June \nth{21}} 
    		& \includegraphics[width=\width]{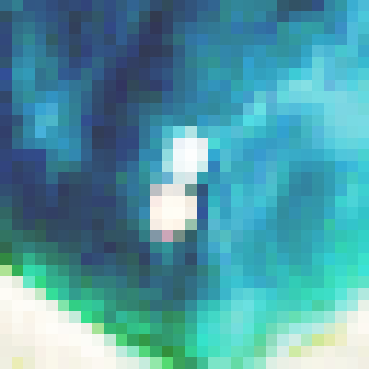} 
    		& \includegraphics[width=\width]{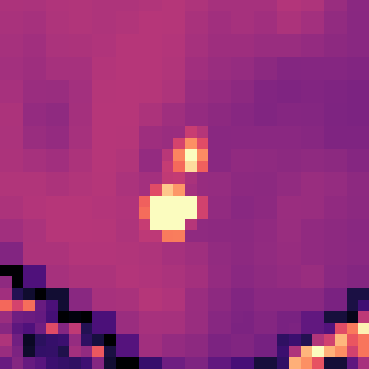} 
    		& \includegraphics[width=\width]{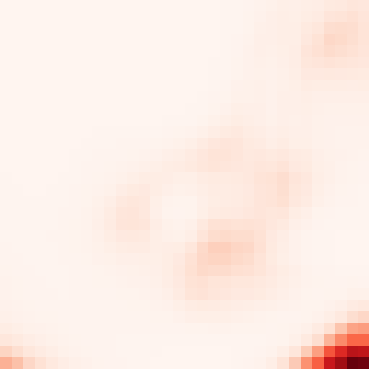} 
    		& \includegraphics[width=\width]{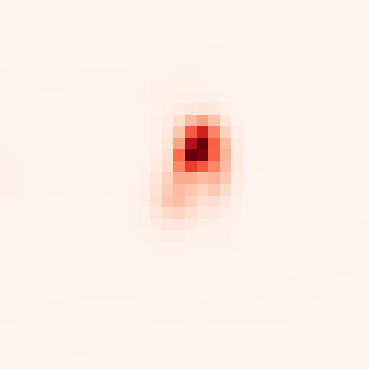} 
    		& \includegraphics[width=\width]{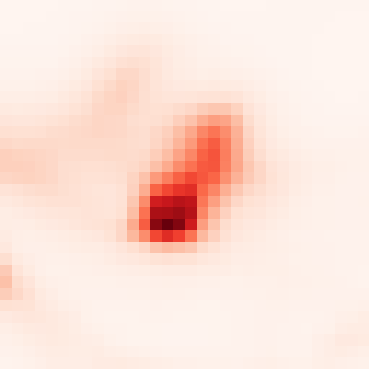} 
    		& \includegraphics[width=\width]{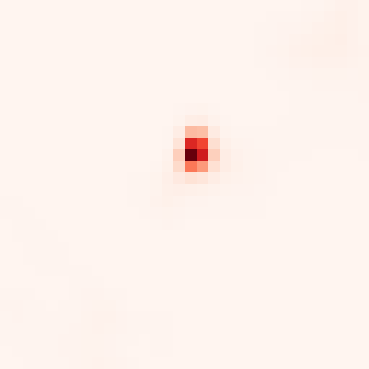} 
    		& \includegraphics[width=\width]{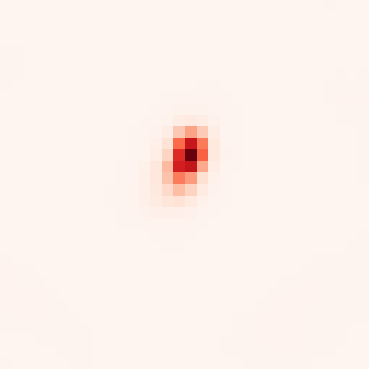} 
    		& \includegraphics[width=\width]{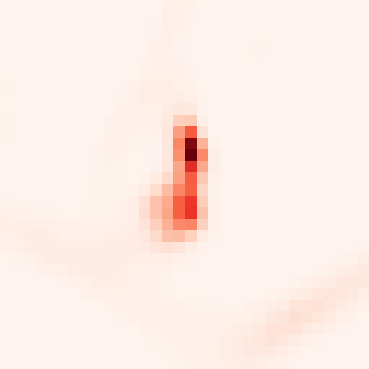} 
    		\\

    		\rotatebox[origin=lB]{90}{July \nth{1}} 
    		& \includegraphics[width=\width]{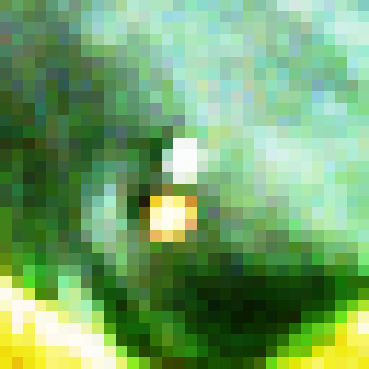} 
    		& \includegraphics[width=\width]{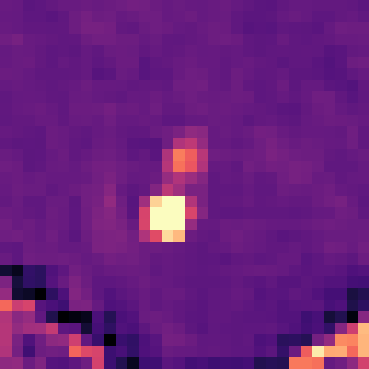} 
    		& \includegraphics[width=\width]{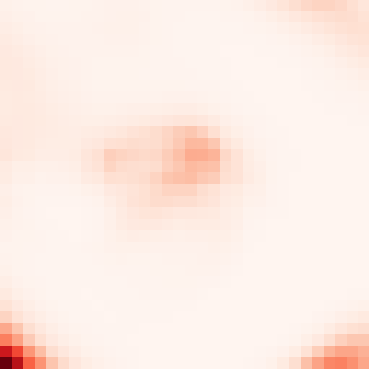} 
    		& \includegraphics[width=\width]{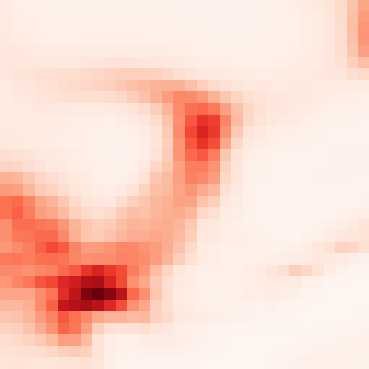} 
    		& \includegraphics[width=\width]{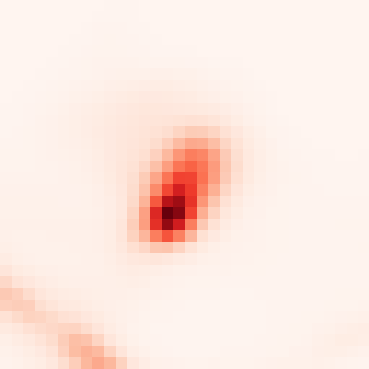} 
    		& \includegraphics[width=\width]{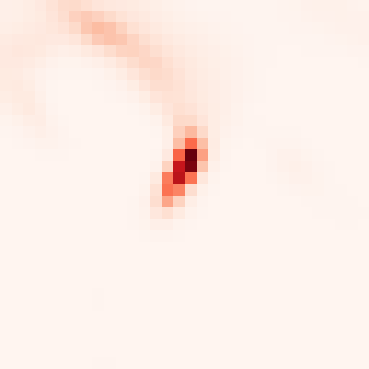} 
    		& \includegraphics[width=\width]{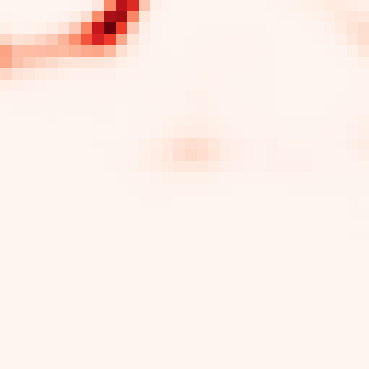} 
    		& \includegraphics[width=\width]{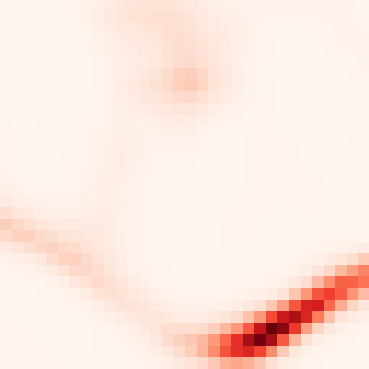} 
    		\\
    		
    		\rotatebox[origin=lB]{90}{July \nth{31} } 
    		& \includegraphics[width=\width]{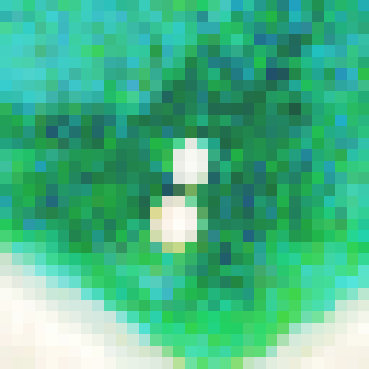} 
    		& \includegraphics[width=\width]{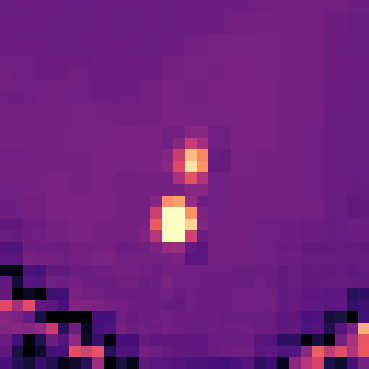}
    		& \includegraphics[width=\width]{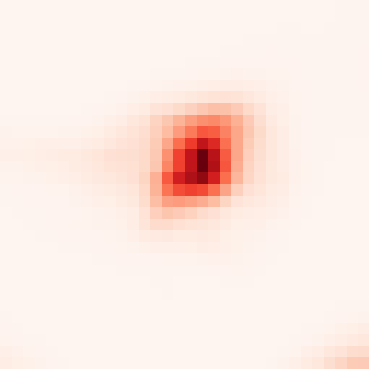} 
    		& \includegraphics[width=\width]{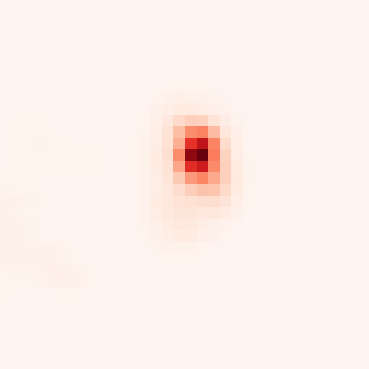} 
    		& \includegraphics[width=\width]{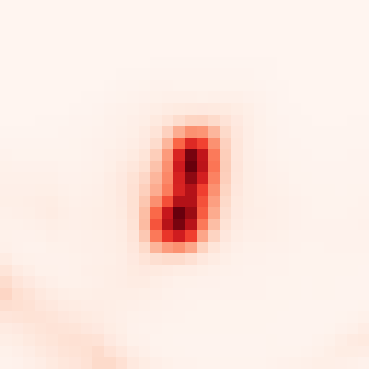} 
    		& \includegraphics[width=\width]{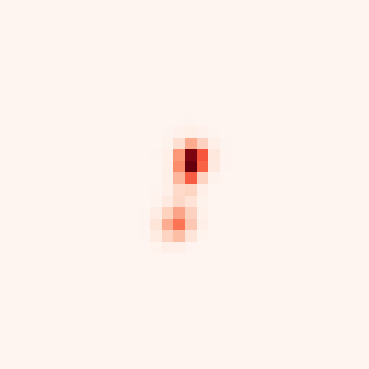} 
    		& \includegraphics[width=\width]{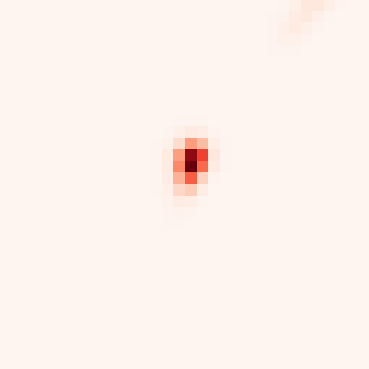} 
    		& \includegraphics[width=\width]{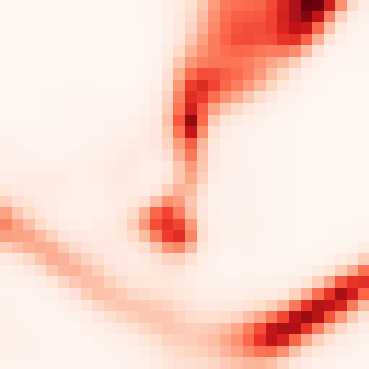} 
    		\\
    		
    		\rotatebox[origin=lB]{90}{Aug. \nth{10}} 
    		& \includegraphics[width=\width]{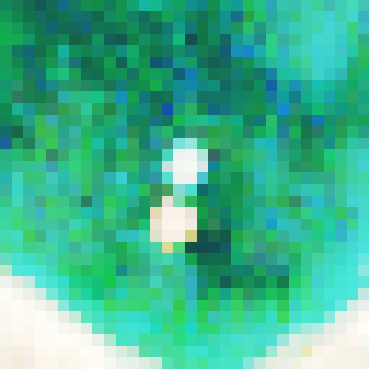} 
    		& \includegraphics[width=\width]{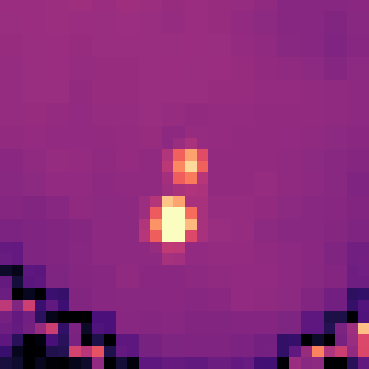}
    		& \includegraphics[width=\width]{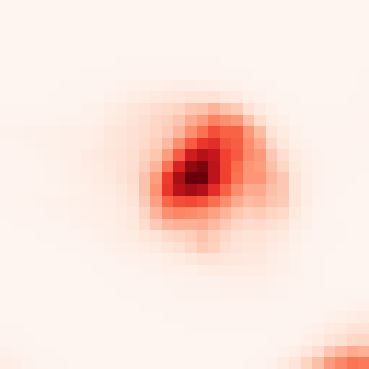} 
    		& \includegraphics[width=\width]{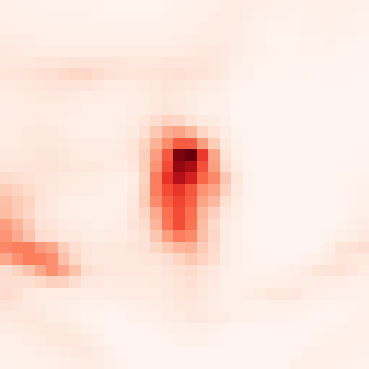} 
    		& \includegraphics[width=\width]{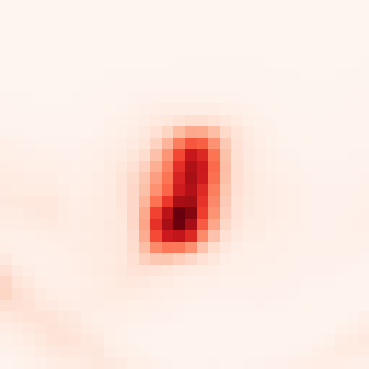} 
    		& \includegraphics[width=\width]{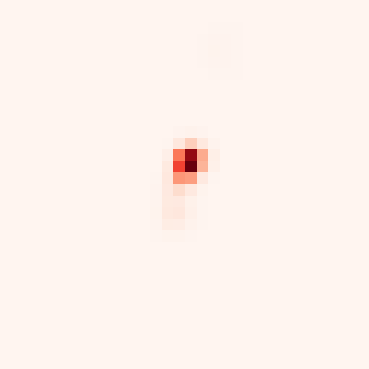} 
    		& \includegraphics[width=\width]{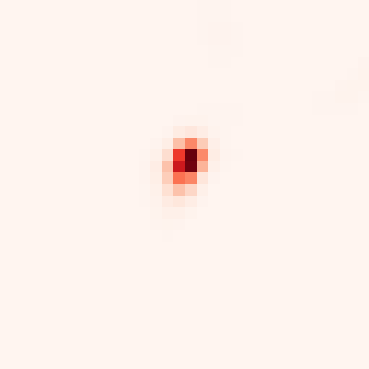} 
    		& \includegraphics[width=\width]{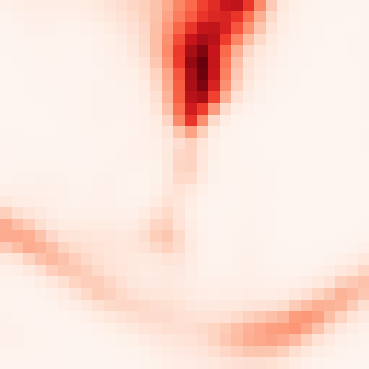} 
    		\\
    	\end{tabular}

	\caption{Classification probabilities for {Sentinel-2} scenes of deployed targets in during the Plastic Litter Projects 2021 \citep{topouzelis2019detection}. All models assign higher probabilities to the deployed targets. Still, only few models detect both targets. Other pixels, such as coastlines, are sometimes assigned a higher marine debris probability. Models trained with the label refinement module tend to predict larger patches with less spatial detail.}
	\label{fig:plp}
\end{figure}

Finally, we compare different \modelname{Unet++} models trained on different initialization seeds, with and without label refinement on images of the Plastic Litter Projects 2021 (\cref{fig:plp}). Most models capture the general location of the deployed targets on all scenes. However, some models (seed 3; no label refinement and seed 2 with label refinement) confuse the coastline and some water areas for marine debris. Seed 1 with label refinement appears to miss the deployed targets on June \nth{21} and July \nth{1}, similarly to the model trained on seed 2 with label refinement on July \nth{1}. Similarly to the previous result, models trained with refined labels predict larger but also less defined patches compared to models trained without. 
This experiment demonstrates the challenges associated with detecting individual objects that span only few pixels. 
However, we would like to highlight that these deployed targets are not representative of the marine debris seen in open waters, on which the models have been trained on. These objects typically form long lines rather than round shapes, and we believe that the difference in geometrical shape, rather than spectral appearance, is a major feature that the deep learning models use for their predictions. 

\subsection{Role of Atmospheric Correction}
\label{sec:confusions}

\begin{figure}
    \centering
    
    \begin{subfigure}{.48\textwidth}
    	\includegraphics[width=\textwidth]{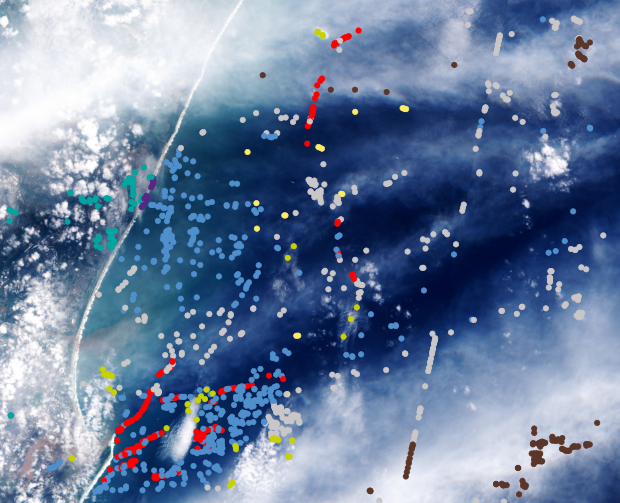}
    	\caption{TOA: top-of-atmosphere  Sentinel-2 scene}
    	\label{fig:confusion:toa}
    \end{subfigure}
	\hfill
    \begin{subfigure}{.48\textwidth}
        \includegraphics[width=\textwidth]{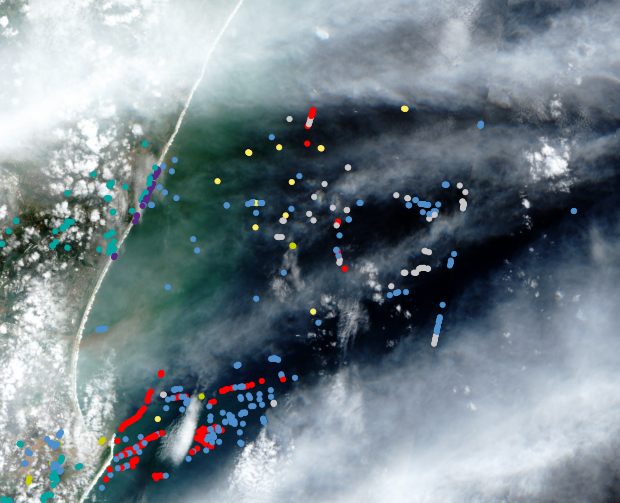}
        \caption{BOA: bottom-of-atmosphere Sentinel-2 scene}
        \label{fig:confusion:boa}
    \end{subfigure}
    \vspace{1em}
    
    \begin{subfigure}{\textwidth}
        \footnotesize 

\makeatletter
\pgfplotsset{
    calculate offset/.code={
        \pgfkeys{/pgf/fpu=true,/pgf/fpu/output format=fixed}
        \pgfmathsetmacro\testmacro{(\pgfplotspointmeta *10^\pgfplots@data@scale@trafo@EXPONENT@y)*\pgfplots@y@veclength)}
        \pgfkeys{/pgf/fpu=false}
    },
    every node near coord/.style={
        /pgfplots/calculate offset,
        yshift=-\testmacro
    },
}

\definecolor{debriscolor}{HTML}{FF0000}
\definecolor{shipscolor}{HTML}{FBEE66}
\definecolor{coastlinecolor}{HTML}{5C2483}
\definecolor{landcolor}{HTML}{00A79F}
\definecolor{cloudscolor}{HTML}{C8D300}
\definecolor{hazedense}{HTML}{5B3428}
\definecolor{hazetrans}{HTML}{CAC7C7}
\definecolor{watercolor}{HTML}{4F8FCC}

 \newcommand{\colorboxes}{
 \begin{tikzpicture}[xscale=1.45]
 \node[fill=debriscolor] at (1,0){};
 \node[fill=hazetrans] at (2,0){};
 \node[fill=hazedense] at (3,0){};
 \node[fill=cloudscolor] at (4,0){};
 \node[fill=shipscolor] at (5,0){};
 \node[fill=landcolor] at (6,0){};
 \node[fill=coastlinecolor] at (7,0){};
 \node[fill=watercolor] at (8,0){};
 \end{tikzpicture}
 }

\pgfplotstableread{
0 136 164  
1 98 600  
2 0 142    
3 14 41  
4 16 18  
5 92 58 
6 26 15 
7 227 446 
}\dataset
    
    \begin{tikzpicture}
\begin{axis}[ybar,
        width=\textwidth,
        height=4cm,
        axis y line*=left,
        axis x line*=bottom,
        font={\sffamily},
        ymin=0,
        ymax=700,        
        ylabel={\#detections},
        xtick=data,
        xticklabels = {
            {\textbf{debris}},
            {\textbf{t. hz.}},
            {\textbf{d. hz.}},
            {\textbf{clouds}},
            {\textbf{ships}},
            {\textbf{land}},
            {\textbf{coast}},
            {\textbf{water}}
        },
        xticklabel style={yshift=-8ex, rotate=0, anchor=north, align=center, yshift=1em},
        major x tick style = {opacity=0},
        minor x tick num = 1,
        minor tick length=2ex,
        every node near coord/.append style={
                anchor=east,
                rotate=90
        }
        ]
\addplot[draw=none,fill=rouge, nodes near coords=TOA] table[x index=0,y index=1] \dataset; 
\addplot[draw=none,fill=leman, nodes near coords=BOA] table[x index=0,y index=2] \dataset; 

\end{axis}

\node at (6.05,-1.6) {\colorboxes};

 \end{tikzpicture}
 
        \caption{number and confusions of detections. Classes evaluated are \classname{marine debris} (the correct class) and confusions with \classname{transparent haze} (t.hz.),  \classname{dense haze} (d.hz), \classname{cummulus clouds} (clouds), \classname{ships}, \classname{land}, \classname{coastline} (coast), and \classname{water}.}
        \label{fig:confusion:barplot}
    \end{subfigure}
    \caption{Analysis of confusions of detections in atmospherically corrected bottom-of-atmosphere (BOA) and not correction top-of-atmosphere (TOA) {Sentinel-2} imagery of the Durban scene. In (a) and (b) panels, detections are colored according to the classes of panel (c).}
    \label{fig:confusion}
    \end{figure}
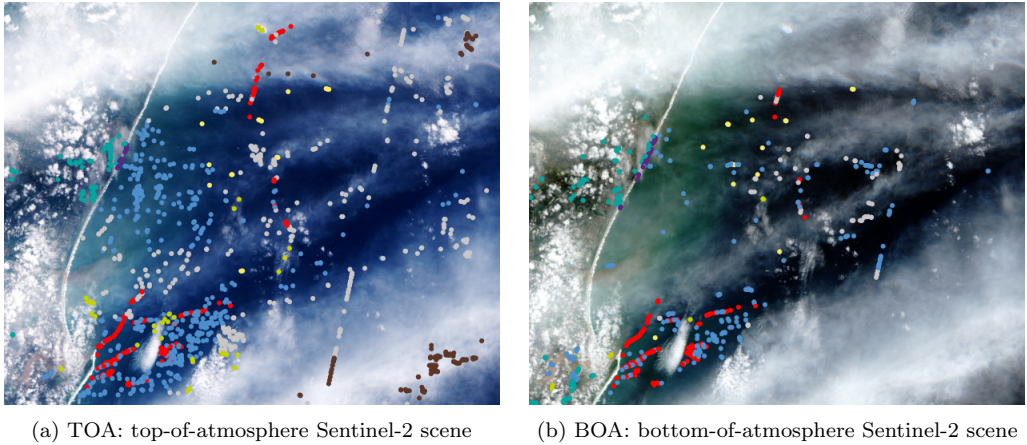
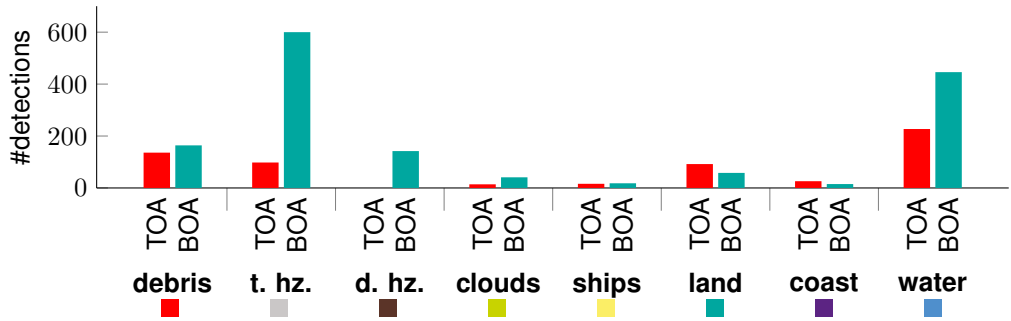

In this experiment, we follow a realistic deployment scenario and predict the entire Durban scene of \SI{3122}{\pixel} $\times$ \SI{3843}{\pixel} with the \modelname{Unet++} model in overlapping \SI{480}{\pixel} $\times$ \SI{480}{\pixel} patches. We then consider pixels predicted with a probability higher than the prediction threshold and treat each local maximum as a marine debris detection. We set a minimum distance of \SI{3}{\pixel} between local maxima to avoid marine debris detections being too close to each other. Furthermore, we compare predictions of the same model using either a top-of-atmosphere (TOA) {Sentinel-2} scene on a bottom-of-atmosphere (BOA) atmospherically corrected {Sentinel-2} scene, to assess the effect of atmospheric correction on the model predictions. 

We show both images alongside the locations of detections (scatter points) in \cref{fig:confusion:boa,fig:confusion:toa}, respectively. The red scatter points indicate correctly detected \classname{marine debris}. Points of other colors indicate false-positives with other classes \classname{transparent haze} (t.hz.),  \classname{dense haze} (d.hz), \classname{cumulus clouds} (clouds), \classname{ships}, \classname{land}, \classname{coastline} (coast), and \classname{water}, alongside \classname{marine debris} (debris). \Cref{fig:confusion:barplot} further shows a quantitative summary of the confusion between classes.
We generally see a comparable number of \classname{marine debris} detected at both BOA (\num{136} detections) and TOA (\num{164} detections) processing levels.
This shows that the classifier is sensitive to marine debris in both top-of-atmosphere and bottom-of-atmosphere satellite imagery.
 However, predictions based on top-of-atmosphere data had more false positive predictions leading to a lower precision. 
 This is especially visible in the \classname{transparent-} (t.hz.) and \classname{dense haze} (d.hz) categories as well, as in water, as shown in the bar plot of \cref{fig:confusion:barplot}. Overall and not shown in the figure: \num{609} objects were detected in the bottom-of-atmosphere (BOA) scene, and \num{1484} objects as \classname{marine debris} in the top-of-atmosphere scene.
For comparison, the \modelname{Unet} trained only on the FloatingObjects dataset of \citep{mifdaletal2020towards} detected \num{20830} objects in the BOA scene and \num{33665} at TOA processing level, which is more than one order of magnitude more false positive predictions compared to the \modelname{Unet++} shown in \cref{fig:confusion}. This demonstrates ever more the importance of compiling larger and more precise training datasets with a rich pool of negative examples that account for objects easily confused with marine debris. It demonstrates the current limitations and general difficulty of detecting marine debris automatically on Sentinel-2 imagery with the current technology. The extreme imbalance between a very low number of {marine debris} pixels (if any) and everything else visible in the {Sentinel-2} scene poses a severe challenge to the automated detection of marine debris. Overall in this experiment, only \num{6448} of \num{11997846} pixels were annotated as {marine debris}, which represents coverage of only 0.05\%. In this circumstance, identifying less than potential \num{1000} objects in a \SI{31}{\kilo\meter} by \SI{38}{\kilo\meter} is an achievement and allows to validate these detections visibly with limited manual effort in practice. This work can be further reduced by additional targeted post-processing by masking clouds, land, and shoreline explicitly, which we consider outside of the scope of this work.

\subsection{Transferability to PlanetScope Resolution}
\label{sec:transferability}

\begin{figure}
\centering
\begin{subfigure}{8cm}
    \includegraphics[width=\textwidth]{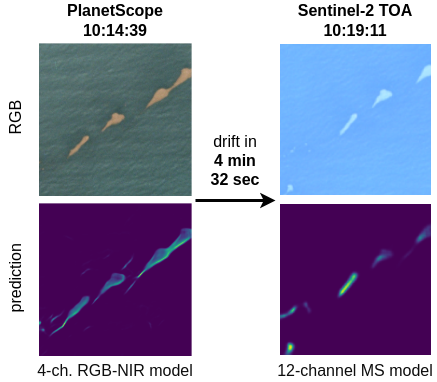}
    \caption{Double-acquisition of {Sentinel-2} and PlanetScope of sargassum patches in Accra (2018-10-30) with 4 minutes 32-second delay.}
    \label{fig:doubleacquisition}
\end{subfigure}

\begin{subfigure}{\textwidth}
    \includegraphics[width=\textwidth]{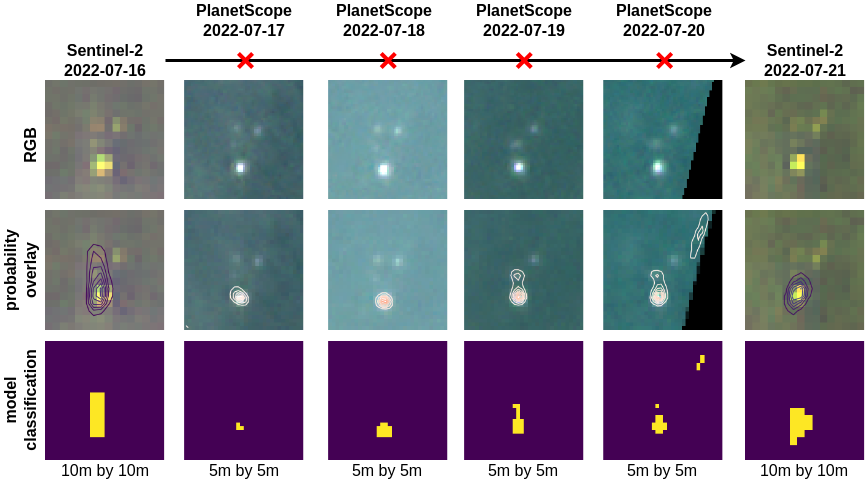}
    \caption{Daily PlanetScope imagery fills the observation gaps of {Sentinel-2} (every 5 days) for the Plastic Litter Project (Island of Lesbos, Greece) where marine plastic and wooden targets were deployed in 2022 \cite{topouzelis2019detection}}
    \label{fig:gapfilling}
\end{subfigure}
\caption{A four-channel RGB+NIR model trained on {Sentinel-2} imagery can classify marine debris in \SI{5}{m} $\times$ \SI{5}{m} downsampled Planetscope images, while being trained on 4-channel {Sentinel-2} imagery. We showcase two use cases. In a) a simultaneous acquisition of S2 and PS in Accra show the drift direction of Sargassum patches. In b) PlanetScope images augment S2 observations in the Plastic Litter Project.}
\label{fig:ps-transferability}
\end{figure}

In this final experiment, we test how well the \modelname{Unet++} model trained on {Sentinel-2} imagery can predict on PlanetScope without being fine-tuned on PlanetScope imagery specifically. 
For this experiment, we had to downsample the PlanetScope imagery from \SI{3}{\meter} to \SI{5}{\meter} as the resolution gap between trained \SI{10}{\meter} resolution and full \SI{3}{\meter} PlanetScope imagery was too large. On the original resolution, the model created artifacts in the predictions, which disappeared at downsampled \SI{5}{\meter} PlanetScope imagery.
For the {Sentinel-2} image, we use the same model with 12 input channels as in the previous experiments. For the 4-channel PlanetScope imagery, we re-trained the \modelname{Unet++} model on the identical {Sentinel-2} training data but removed all spectral bands except B2, B3, B4, and B8 for RGB+NIR. This 4-channel model achieves a slightly worse validation accuracy (0.01-0.03 in \metric{f-score}) than the 12-channel model. This slight decrease in accuracy also indicates that the four high-resolution \SI{10}{\meter} bands are the most informative for marine debris detection, which is reasonable given the small size of debris and previous literature (\cite{biermann2020finding}).

We consider two use cases in \cref{fig:ps-transferability}, where PlanetScope imagery complements {Sentinel-2}.
\begin{itemize}
	\item First, double acquisitions of {Sentinel-2} and PlanetScope during the same day can be used to determine the debris's short-term surface drift direction. It shows one PlanetScope with a corresponding {Sentinel-2} image over Accra, Ghana, on \nth{30} of October 2018, with four minutes and 32 seconds time difference. Both models detected marine debris, as visible in the probability map. 
	\item Second, daily PlanetScope imagery can be used to gap-fill the periods in which the weekly {Sentinel-2} imagery is unavailable. This is demonstrated in \cref{fig:gapfilling}, where the deployed targets from the Plastic Litter Project 2022 are predicted from {Sentinel-2} and PlanetScope imagery with the \modelname{Unet++} model. The {Sentinel-2} images are available only on July \nth{16} and \nth{21}. Daily PlanetScope imagery can fill this temporal gap and enable continuous monitoring of the deployed targets at a higher spatial-, but lower spectral resolution. We can see that the 4-channel model successfully predicts marine debris for the rectangular \SI{5}{\meter} $\times$ \SI{5}{\meter} inflatable PVC target deployed during the Plastic Litter Project. The two circular (\SI{7}{m} diameter) HDPE-mesh targets are not detected. 
	
\end{itemize}

Thanks to these two examples, we emphasize that the \modelname{UNet++} model in our Marine Debris Detector trained on {Sentinel-2} imagery worked with PlanetScope images without explicitly having seen annotated PlanetScope imagery. 
This highlights the broader applicability of the \modelname{Unet++} model on both satellite modality and the synergy between PlanetScope and {Sentinel-2} satellite constellations for marine debris detection.

\section{Discussion}
\label{sec:discussion_and_conclusion}

\new{This work presented and evaluated a training strategy including a dataset, targeted negative sampling and a segmentation model} to automatically identify marine debris \new{of human or natural origins} with readily available {Sentinel-2} imagery. 
Our main contribution is the aggregation and harmonization of all annotated Sentinel-2 data for marine debris detection available today. We designed a sampling rule to gather a large number of diverse negative examples and a refinement module to automatically improve hand-annotations present in current datasets, which yields a combined training dataset in which deep learning models achieve the best results across different model architectures. 
The model performances were compared quantitatively and qualitatively on evaluation scenes where the visible marine debris in these scenes is highly likely to contain plastic pollutants. 
\new{The performance improvements observed are consistent across datasets and model settings.
They highlight the importance of designing good datasets for the tasks at hand and prove the necessity to collect, aggregate and further refine globally distributed datasets of marine debris in future research.}


\new{\textbf{Role of atmospheric correction}}. Atmospheric correction with \new{Sen2Cor} has proven beneficial in reducing the number of false positive examples and improving precision. Still, the detector remained sensitive to marine debris also with top-of-atmosphere data, \new{which highlights the the sensitivity of the model to marine debris.}
We believe that reliably detecting marine debris from available satellite data is within reach with more annotation and targeted post-processing, such as automatic masking of clouds, land, and shoreline, which we considered beyond the scope of this work.
\new{In this work, we trained the detector with Sentinel-2 images of both top-of-atmosphere (L1C-level) and bottom-of-atmosphere (L2A-level with the Sen2Cor algorithm) to ensure that the final model is capable of detecting marine debris from Sentinel-2 imagery at different processing levels. However, further atmospheric correction specific for coastal and aquatic environments, as with the ACOLITE algorithm \citep{vanhellemont2016acolite}, is likely to improve the detection accuracy further.}
 
\new{\textbf{Marine debris as proxy for marine litter.}} The detection of marine debris remains a proxy objective targeted toward the long-term goal of enabling continuous monitoring of marine \new{litter including plastics and other anthropogenic pollutants} from \new{medium-resolution} satellite data.
Here, automatically establishing the link between detected marine debris and marine pollution is a key question to be addressed in the future. Similar to related work \citep{biermann2020finding}, we analyzed social media (Durban scene) and in-situ studies (Accra scene) on a case-by-case basis to deduce that marine plastics are present in marine debris visible in the satellite scenes. 
Automating this connection remains a challenge that may require integrating in-situ knowledge (citizen science, or river monitoring) or a targeted acquisition and analysis of high-resolution imagery.
Studies \citep{cozar2021,ruiz2020litter} have demonstrated that plastics are present in marine debris by on-site ship-based collection. This establishes that marine debris detection is a suitable, yet rough, proxy for plastic pollution mapping.
\new{Ongoing research \citep{HU2021112414,hu2022spectral,CIAPPA2021112457} in this field demonstrates that distinguishing anthropogenic marine litter from natural types of debris using only features is possible, but remains challenging and is largely unsolved today.
Our work concentrated on the prior step of automating the detection of generic marine debris at a large scale largely based on their geometric shape, which can be seen as a first step preceeding the aforementioned litter types characterization.}

\new{\textbf{Relevance for of Algae and Sargassum Detection}}. While the evaluation datasets in our work aimed to measure the detector's sensitivity to marine litter, we see that the model is also sensitive to detections of floating algae patches and sargassum. This sensitivity is inherently connected to the annotations in the training dataset that were made by visually inspecting the Floating Debris Index \citep{biermann2020finding} that is derived from the Floating Algae Index \citep{hu2009novel}. Hence, exploration and modification of the training framework presented in this work and initialization from model weights and fine-tuning towards detecting patches of algae and sargassum would be an interesting follow-up work in an active research field \citep{wang2021satellite,cuevas2018satellite}.

\new{\textbf{Transfer to other satellite products}}. The synergy of Sentinel-2 with daily available PlanetScope (or other high-resolution imagery) is particularly suitable for further analysis of detected debris and establishing a connection to marine litter. Large-scale monitoring with commercial high-resolution imagery may be infeasible due to the high image acquisition costs. However, selecting a few images with PlanetScope in locations where a Sentinel-2 detector has identified potential marine debris appears feasible. We explored this transferability in \cref{sec:transferability} where a model trained on 4-channel Sentinel-2 imagery was still sensitive to marine debris in (downsampled) planet scope data. Targeted model training on annotated PlanetScope data will likely improve this performance further, which we leave for future work.

\new{\textbf{Spatial and spectral features}}. A further direction to be explored is the heterogeneous composition of objects in marine debris, which varies depending on circumstances (e.g., Flood event in Durban) or the general pollution of the area (Accra scene).
This heterogeneity in spectral response further emphasizes the importance and descriptiveness of the shape and geometry in marine debris, which often form elongated lines due to oceanic processes, such as windrows and waterfronts. Further, the geometry of objects is also a suitable descriptor to exclude a variety of negatives, such as ships, clouds, coastline, and wakes, that can have similar spectral responses (e.g., a high FDI index) to marine debris but are distinguishing from marine debris by spatial context. In particular, convolutional neural networks are suitable to learn these patterns in their filter banks if they are trained with large annotated datasets with a diverse set of negative examples. 

\new{
\section{Conclusion}

Remote sensing combined with current machine learning frameworks has the potential to become an efficient and reliable tool to monitor large marine areas \citep{hanke2013guidance}. Still, the data quality used to learn detection models is paramount. We are confident that automated detection of marine debris with satellite remote sensing imagery will provide a repeatable low-cost technology to detect and quantify the level of marine pollution on our planet.}
Automated detection and quantification will be necessary to inform clean-up operations and measure local policy decisions' effect. Identifying and quantifying pollution hotspots and addressing the drivers and sources are crucial to create a cleaner environment to plant, animal, and human life in a sustainable future. 
Still, further efforts are needed in data collection and on-site validation to build models that can reliably estimate the level of marine pollution from readily available satellite data in a completely automated way.
In this research, we made a step toward automated satellite-based monitoring of marine pollution via detecting marine debris in coastal waters and providing model weights and training scripts in a dedicated package \footnote{The source code and data: \url{https://github.com/MarcCoru/marinedebrisdetector}}. We hope this work helps accelerate the progress toward large-scale marine litter monitoring within the canon of trans-disciplinary machine learning, remote sensing, and marine science research.

\bibliographystyle{elsarticle-harv}
\bibliography{references.bib}

\appendix

\end{document}